\documentclass{article}
\PassOptionsToPackage{numbers,sort&compress}{natbib}
\usepackage[main,final]{neurips_2026}

\usepackage[utf8]{inputenc}
\usepackage{tabularx}
\usepackage{multirow}
\newcolumntype{C}{>{\centering\arraybackslash}X}
\usepackage[T1]{fontenc}
\usepackage{graphicx}
\usepackage{hyperref}
\usepackage{url}
\usepackage{booktabs}
\usepackage{amsfonts}
\usepackage{amsmath}
\usepackage{placeins}
\usepackage{amssymb}
\usepackage{wrapfig}
\usepackage{booktabs}
\usepackage[table]{xcolor}
\usepackage{multirow}
\usepackage{makecell}
\usepackage[table]{xcolor}
\usepackage{array}
\usepackage{graphicx}
\usepackage{nicefrac}
\usepackage{microtype}
\usepackage{xcolor}
\usepackage{multirow}
\usepackage{makecell}
\usepackage{caption}
\usepackage{booktabs}
\usepackage{enumitem}
\usepackage{multirow}
\usepackage{makecell}
\usepackage{adjustbox}

\usepackage[table]{xcolor}
\usepackage[most]{tcolorbox}
\tcbuselibrary{breakable}
\usepackage{ragged2e}

\definecolor{RVBase}{RGB}{210,95,20}
\definecolor{RVLLM}{RGB}{35,135,70}
\definecolor{RVVLM}{RGB}{45,95,180}

\newcommand{\cmark}{\textcolor{RVLLM}{\ensuremath{\checkmark}}}
\newcommand{\xmark}{\textcolor{red}{\ensuremath{\times}}}

\newtcolorbox{caseheader}[1]{
  breakable,
  enhanced,
  colback=white,
  colframe=black!30,
  boxrule=0.45pt,
  arc=1.2pt,
  left=6pt,
  right=6pt,
  top=5pt,
  bottom=5pt,
  title={#1},
  fonttitle=\bfseries,
  coltitle=black,
  colbacktitle=black!4,
  before skip=0.8em,
  after skip=0.45em
}

\newtcolorbox{traceblock}[3]{
  breakable,
  enhanced,
  colback=#1!6,
  colframe=#1!35,
  boxrule=0.35pt,
  arc=1.0pt,
  left=6pt,
  right=6pt,
  top=5pt,
  bottom=5pt,
  title={\textbf{#2}\hfill \textbf{Prediction: #3}},
  fonttitle=\bfseries,
  coltitle=black,
  colbacktitle=#1!13,
  before skip=0.35em,
  after skip=0.35em
}

\title{Can Linguistic Reasoning Vectors Enhance Multimodal Reasoning Ability?}

\author{
\normalfont
\textbf{Ziyi Wang}$^{1}$ \quad
\textbf{Li Li}$^{1}$ \quad
\textbf{Aolin Zhou}$^{1}$ \quad
\textbf{Yankun Shen}$^{1}$ \\
\textbf{Chonghan Liu}$^{2}$ \quad
\textbf{Shuxia Lin}$^{1,*}$ \quad
\textbf{Xu Yang}$^{1}$ \\[4pt]
$^{1}$ School of Computer Science \& Engineering, Southeast University, China \\
$^{2}$ University of California, Los Angeles, USA \\[4pt]
\texttt{220256460@seu.edu.cn} \qquad
\texttt{shuxia.mumu@gmail.com} \\[2pt]
}

\begin{document}

\maketitle

\begin{abstract}
% Most Vision-Language Models (VLMs) are built by extending pretrained Large Language Models (LLMs) with visual modules and multimodal alignment, yet it remains unclear how this multimodal scaling affects the language-side reasoning ability encoded in the base LLM. If this ability remains usable after alignment, VLM reasoning may be improved by reusing it from the base LLM without retraining the VLM backbone. 
Most Vision-Language Models (VLMs) are built by extending pretrained Large Language Models (LLMs) with visual modules and multimodal alignment. However, this multimodal scaling often degrades the language-side reasoning ability originally encoded in the base LLM. While the base LLM retains usable reasoning after scaling, the aligned VLM itself cannot reliably access this ability. Therefore, recovering the degraded reasoning capability in VLMs would benefit more from seeking help from the base LLM than from the VLM alone.
Motivated by this, we propose \textbf{LIFT} (\textbf{L}anguage-side reason\textbf{i}ng \textbf{F}acilitation and \textbf{T}ransfer), a lightweight vector-intervention method 
% for studying and improving VLM reasoning. 
that transfers reasoning capability from the base LLM to the VLM without retraining the backbone. 
LIFT defines \textbf{Reasoning Vectors} as answer-token hidden-state differences between a Reasoner path with an explicit reasoning trace and a Solver path without it, and injects these vectors into language-side activations of the target VLM. LIFT further supports learnable vector adaptation while keeping the VLM backbone frozen. 
We evaluate LIFT on two VLMs across six reasoning benchmarks, comparing Reasoning Vectors extracted from the base LLM and from the aligned VLM under matched protocols.
% LLM-derived vectors provide stronger and more consistent improvements than VLM-derived vectors. 
% By transferring and adapting base-LLM Reasoning Vectors, LIFT helps recover this ability through lightweight language-side interventions. 
Results show that LLM-derived vectors consistently outperform VLM-derived vectors, confirming that the base LLM is a more effective source for recovering reasoning. LIFT partially recovers degraded reasoning through lightweight language-side interventions.
Further analyses show that Reasoning Vectors influence intermediate reasoning behavior rather than merely altering final answers. The source code is publicly available at
\url{https://github.com/zitai-wang/lift}.
\end{abstract}

\section{Introduction}

 Most Vision-Language Models (VLMs) are built by extending pretrained Large Language Models (LLMs) with visual modules. A typical architecture employs an LLM backbone, a vision encoder, and a cross-modal projector, achieving multimodal alignment through instruction tuning~\citep{alayrac2022flamingo,li2023blip,liu2023visual,zhu2023minigpt,dai2023instructblip,huang2023language}. This modular design is widely adopted because pretrained LLMs already provide strong language and reasoning capabilities. This raises the possibility that part of a VLM’s reasoning ability is inherited from the underlying LLM, rather than acquired solely through multimodal alignment. However, multimodal scaling may change the behavior and internal representations of the original language model~\citep{ratzlaff2025training,khayatan2025analyzing}. This leads to our central question: whether reasoning representations formed in the base LLM can still be used to intervene on the corresponding VLM.

As a motivating observation, we compare the base LLM and its corresponding VLM on the same pure-text reasoning benchmarks without visual inputs. We find that the base LLM can sometimes outperform the VLM, even though the VLM shares the same language backbone and is further trained with multimodal data. The gap also appears in the reasoning process. In many failure cases, the VLM starts in the right direction but makes mistakes in later reasoning steps. It may copy a number incorrectly, change an intermediate variable, drop a condition, or produce an answer mismatched with the question target. These failures suggest that the relevant reasoning ability remains usable, but becomes less stable during multi-step generation.

Existing studies have made substantial progress in multimodal reasoning by directly improving the target VLM, including chain-of-thought supervision~\citep{zhang2023multimodal,wei2022chain,wang2022self,zhou2022least}, rationale-rich instruction tuning~\citep{guo2025mammoth}, verification-based reasoning~\citep{sun2025mm}, and reasoning-oriented reinforcement learning or post-training~\citep{yang2025r1,huang2025vision}. These methods primarily optimize the VLM’s external behavior, and do not directly examine whether the reasoning-induced representation shifts learned before multimodal alignment remain recoverable and transferable after alignment. However, since the base LLM already has language-side reasoning ability before multimodal alignment, VLM reasoning may be improved by reusing these representations rather than relying exclusively on heavy VLM retraining. Motivated by this view, we study whether the reasoning-induced hidden-state change in the base LLM can be extracted and reused as a lightweight intervention in the VLM, without updating parameters.

\begin{figure}[t]
    \centering
    \includegraphics[width=0.9\linewidth]{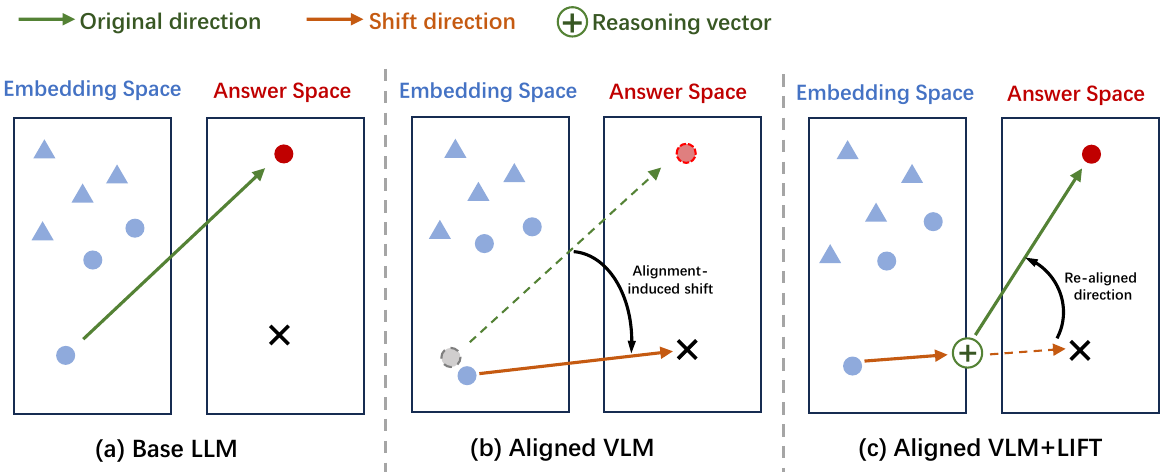}
    \caption{
    Motivation of LIFT. Multimodal alignment may change language-side reasoning directions in the VLM. LIFT tests whether Reasoning Vectors from the base LLM can be reused as lightweight interventions in the aligned VLM.
    }
    \vspace{-2.5em}
    \label{fig:intro_lift}
\end{figure}

Recent work on task vectors suggests that task-relevant knowledge can be summarized as compact activation or parameter differences~\citep{ilharco2022editing,panickssery2023steering}. Such vectors can steer model behavior without modifying the full model, providing a natural tool for studying VLM reasoning after multimodal alignment. If alignment changes how the VLM accesses language-side reasoning states, then a reasoning-induced hidden-state shift extracted from the base LLM may serve as a useful intervention direction. Building on this view, we propose \textbf{LIFT} (\textbf{\underline{L}}anguage-side reason\textbf{\underline{i}}ng \textbf{\underline{F}}acilitation and \textbf{\underline{T}}ransfer), a lightweight vector-intervention method for VLM reasoning. Given a question, a reasoning trace, and a final answer, we construct two teacher-forced paths. The Reasoner path takes the question--reasoning--answer context, while the Solver path takes only the question--answer context. Because the two paths share the same question and final answer, we use their hidden-state difference on answer tokens as an approximation of the reasoning-induced shift in answer representations. We define this difference as the \textbf{Reasoning Vector}. LIFT injects these vectors into the target VLM as lightweight interventions.

We instantiate LIFT with three extraction sources. LLM-derived extracts Reasoning Vectors from the base LLM using text-only reasoning data, capturing reasoning-induced directions before multimodal scaling.  VLM-derived extracts vectors from the target VLM using the same text-only data, examining how these directions appear after multimodal alignment. VLM-MM extracts vectors from the target VLM using multimodal reasoning data, where images are included as context but the vectors are still computed from language-side answer representations.

Across Qwen2.5-VL-7B-Instruct~\citep{bai2025qwen25vltechnicalreport} and InternVL2.5-8B~\citep{chen2024expanding}, Reasoning Vectors extracted from the corresponding base LLMs consistently improve VLM performance and outperform vectors extracted from the VLMs. Learnable adaptation further improves performance, especially when initialized from the base-LLM vectors. These results suggest that language-side reasoning representations remain useful after multimodal scaling, but may become less directly accessible in the aligned VLM.

Overall, our main contributions are as follows:\begin{itemize}[leftmargin=1.2em, itemsep=0.2em, topsep=0.2em, partopsep=0pt, parsep=0pt]

\item We introduce LIFT, a lightweight vector-intervention method for VLM reasoning. LIFT provides a framework for extracting and transferring language-side Reasoning Vectors from base LLMs to their corresponding VLMs after multimodal scaling.

\item We show that Reasoning Vectors extracted from the base LLM provide stronger interventions than those from the corresponding VLM. 
% This suggests that language-side reasoning representations remain usable after multimodal scaling, but can become harder to recover from the aligned VLM.
This suggests that language-side reasoning ability remains usable after multimodal scaling, and that the corresponding base LLM is a more effective source for recovering degraded reasoning in the aligned VLM.
Learnable vector adaptation further recovers access to this reasoning ability in the target VLM.

\item We analyze how  Reasoning Vectors affect VLM reasoning through source comparisons and layer-wise analysis. These show that Reasoning Vectors influence the reasoning process rather than merely changing the final answer, often helping the model stay aligned with the question target.
\end{itemize}

\section{Related Work}
\label{sec:related_work}

\noindent\textbf{Task and activation vectors.}
Task and activation vectors provide compact representations in parameter or representation space for steering model behaviors~\citep{subramani2022extracting,turner2023steering,zou2023representation,panickssery2023steering,khayatan2025analyzing}. Prior work constructs such vectors either from parameter differences between fine-tuned and pretrained models~\citep{ilharco2022editing,ortiz2023task,li2025task} or from activation differences induced by different prompts, contexts, or demonstrations~\citep{liu2023context,todd2023function,he2021towards,jiang2025mimicincontextlearningmultimodal}. In language models, these vectors have been used for parameter-efficient adaptation, model editing, in-context learning, and controllable activation steering~\citep{yang2025unifying}. More recently, \citep{zhang2025uncovering} studied latent
chain-of-thought vectors, while
\citep{li2025cotvectorstransferringprobing} investigated
the transfer and probing of reasoning mechanisms using
CoT vectors. These studies show that injecting reasoning-related
vectors into hidden states can improve multi-step reasoning.
Our work is closest to activation-level steering, but the goal
is different. Prior work mainly studies how activation directions
represent or control behaviors within language models, whereas
LIFT uses reasoning-conditioned activation differences as a
probe for multimodal scaling. We study whether Reasoning Vectors
extracted from the base LLM remain useful after multimodal
scaling, and compare them with vectors extracted directly from
the corresponding VLM.

\noindent\textbf{Reasoning enhancement and transfer in VLMs.}
Prior work has improved VLM reasoning along several directions. At inference time, methods often guide generation through tree-of-thought search~\citep{yao2024mulberry,yao2023tree}, question decomposition~\citep{zhang2024visual}, verification~\citep{sun2025mm}, or reward-guided decoding~\citep{manas2025controlling}. Regarding training, existing studies introduce multimodal CoT supervision or rationale-rich instruction tuning~\citep{zhang2023multimodal,guo2025mammoth,thawakar2025llamav,shao2024visual,xu2025llava,zhao2026thinking}, while recent reasoning-oriented post-training approaches utilize reinforcement learning to further refine performance~\citep{yang2025r1,huang2025vision,chen2026omnireasonerthinkinglongaudiovideo}. Other studies investigate the cross-modal transfer of reasoning capabilities, such as leveraging textual CoT data to compensate for scarce visual annotations~\citep{du2025virgo}, steering internal reasoning states~\citep{luo2025ursa}, or merging LLM and VLM parameters~\citep{chen2025bring}. These advancements demonstrate the vast potential for enhancing VLM reasoning through optimized training and decoding paradigms. Rather than introducing another training or decoding strategy, our focus is to use language-side Reasoning Vectors as probes and interventions for studying whether reasoning representations formed in the base LLM remain recoverable in the aligned VLM after multimodal scaling. We therefore compare vectors extracted from the base LLM and from the corresponding VLM under the same Reasoner–Solver extraction and intervention protocol.

\section{Methodology}
\label{sec:methodology}

\begin{figure}[t]
    \centering
    \includegraphics[width=0.90\linewidth]{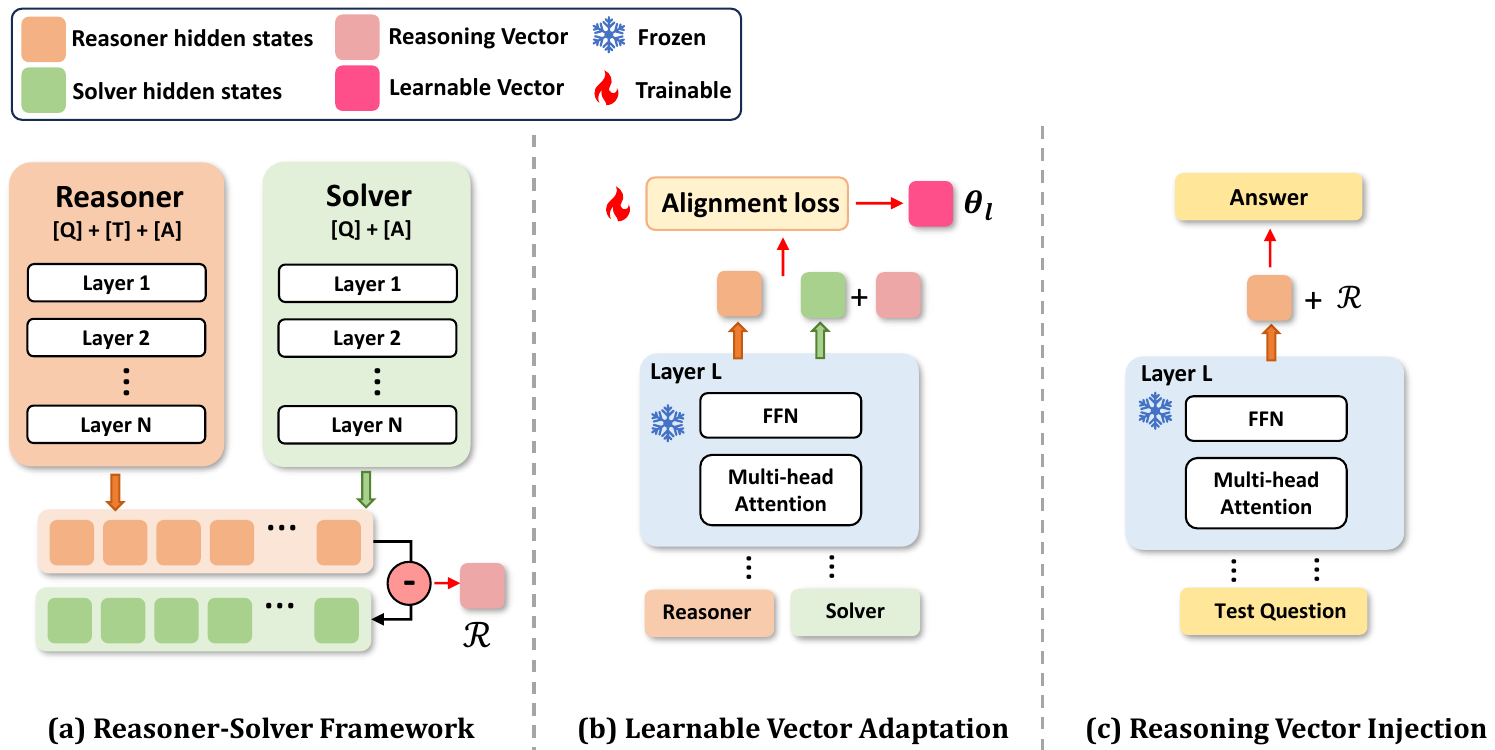}
\caption{
Overview of LIFT. (a) For each layer, LIFT computes hidden-state differences between the Reasoner and Solver paths and averages these differences to obtain the Reasoning Vector. (b) Learnable adaptation optimizes only the injected vector parameters while keeping the VLM backbone frozen. (c) At inference, vectors are injected into selected VLM layers.
}
    \label{fig:method}
\end{figure}

We propose \textbf{LIFT} (\textbf{\underline{L}}anguage-side reason\textbf{\underline{i}}ng \textbf{\underline{F}}acilitation and \textbf{\underline{T}}ransfer), a lightweight vector-intervention method for studying and improving VLM reasoning with language-side Reasoning Vectors. Rather than updating the VLM backbone, LIFT defines a Reasoning Vector as the answer-token hidden-state difference between Reasoner and Solver paths, using it as an approximation of the reasoning-induced shift in answer representations. The resulting vectors are used as interventions in selected language-model layers. We study LIFT in two settings: static vector transfer, where Reasoning Vectors from different sources are directly injected into the target VLM, and learnable vector adaptation, where only activation-level vector parameters are optimized. Figure~\ref{fig:method} provides an overview of the overall pipeline. Section~\ref{subsec:reasoning_vector} formulates Reasoning Vectors through a Reasoner--Solver comparison. Section~\ref{subsec:vector_sources} describes vector sources and extraction, while Sections~\ref{subsec:learnable_vector} and~\ref{subsec:injection_layer_selection} detail learnable vector adaptation and inference-time injection, respectively.

\subsection{Reasoning Vector Formulation}
\label{subsec:reasoning_vector}

Given a question $Q$, a reasoning trace $T$, and a final answer $A$, we construct two teacher-forced forward paths using the same model. The \textbf{Reasoner} path receives the question–reasoning–answer context, denoted as $\mathcal{C}_{R}=(Q,T,A)$, while the \textbf{Solver} path receives the question–answer context, denoted as $\mathcal{C}_{S}=(Q,A)$. Hidden states are read only from aligned answer-token positions in the final answer span, while tokens from the question and reasoning trace are excluded.

Let $\boldsymbol{h}_l(a;\mathcal{C})$ denote the hidden representation of answer token $a$ at layer $l$ under context $\mathcal{C}$. For each answer token $a$, we first compute the token-wise Reasoner–Solver hidden-state difference:
\begin{equation}
\boldsymbol{\delta}_l(a)
=
\boldsymbol{h}_l(a;\mathcal{C}_{R})
-
\boldsymbol{h}_l(a;\mathcal{C}_{S}),
\quad a\in\mathcal{A}.
\label{eq:token_reasoning_vector}
\end{equation}
This difference measures how the answer-token representation changes when the reasoning trace is included in the context. Since the two paths share the same question and final answer, the difference provides an approximation of the reasoning-induced shift in answer representations.

For a support set $\mathcal{D}=\{(q_i,t_i,a_i)\}_{i=1}^{N}$, we average the token-wise differences over the final answer span to obtain a task-level Reasoning Vector at layer $l$:
\begin{equation}
\boldsymbol{r}_l
=
\frac{1}{N}
\sum_{i=1}^{N}
\frac{1}{|\mathcal{A}_i|}
\sum_{a\in\mathcal{A}_i}
\left[
\boldsymbol{h}_l(a;\mathcal{C}_{R,i})
-
\boldsymbol{h}_l(a;\mathcal{C}_{S,i})
\right],
\label{eq:task_reasoning_vector}
\end{equation}
where $\mathcal{C}_{R,i}=(q_i,t_i,a_i)$ and $\mathcal{C}_{S,i}=(q_i,a_i)$. We compute this average independently at each language-model layer. 
Thus, Eq.~\eqref{eq:task_reasoning_vector} defines a task-level Reasoning Vector $\boldsymbol{r}_l$ for layer $l$, and the full LIFT representation is the layer-wise collection $\mathcal{R}=\{\boldsymbol{r}_l\}_{l=1}^{L}$.
% At inference time, LIFT adds the vector to the hidden state of the selected language-model layer, with $\mu$ controlling the intervention strength. We describe the detailed injection procedure in Section~\ref{subsec:injection_layer_selection}.
% \begin{equation}
% \boldsymbol{h}_l(a;\mathcal{C}_{R})
% \approx
% \boldsymbol{h}_l(a;\mathcal{C}_{S})
% +
% \mu\boldsymbol{r}_l,
% \label{eq:vector_intervention_view}
% \end{equation}

\subsection{Reasoning Vector Sources and Extraction}
\label{subsec:vector_sources}

For each extraction source, we build the support set from samples that the source model answers correctly. We then apply the same Reasoner–Solver protocol to compute its Reasoning Vectors. This allows us to compare reasoning directions obtained from each model’s own successful behavior, rather than requiring all models to follow the same reasoning format. 

We consider three vector sources. LLM-derived vectors are extracted from the base LLM using text-only reasoning support sets, capturing reasoning-induced directions before multimodal scaling. VLM-derived vectors are extracted from the target VLM's language backbone using text-only reasoning support sets, testing whether useful language-side reasoning directions can still be obtained after multimodal scaling. VLM-MM vectors are extracted from the target VLM using multimodal reasoning support sets, testing whether multimodal inputs provide directions better aligned with the target VLM's reasoning process.

Here, language-side refers to the extraction and intervention location: all Reasoning Vectors are computed from answer-token hidden states and injected into the language backbone, rather than the vision encoder. Thus, LLM-derived and VLM-derived use text-only support sets, whereas VLM-MM uses multimodal support sets.

\subsection{Learnable Vector Adaptation}
\label{subsec:learnable_vector}

Static vector transfer tests whether an extracted Reasoning Vector can be directly reused by the target VLM. To better align the vector with the target representation space while keeping the VLM backbone frozen, we introduce learnable vector adaptation. We initialize the learnable vector parameter using the extracted Reasoning Vector at the selected adaptation layer, and optimize only this parameter during adaptation.

% Frozen vector transfer tests whether an extracted Reasoning Vector can be directly reused by the target VLM. To further align the vector with the target representation space without updating the VLM backbone~\citep{li2021prefix,hu2021loralowrankadaptationlarge}, we introduce \textbf{learnable vector adaptation}. For selected trainable layers $\mathcal{L}_{\mathrm{train}}$, we initialize trainable vector parameters from extracted vectors:
% \begin{equation}
% \boldsymbol{\theta}_l \leftarrow \mathcal{R},
% \quad l\in\mathcal{L}_{\mathrm{train}}.
% \label{eq:learnable_init}
% \end{equation}
% During adaptation, the VLM backbone remains frozen and only $\{\boldsymbol{\theta}_l\}_{l\in\mathcal{L}_{\mathrm{train}}}$ are updated.
During adaptation, we use the support set to obtain Reasoner representations and intervened Solver representations. For the Solver path, the learnable vector is added to the hidden states at the selected adaptation layer:
% For each support example, the Reasoner takes $(Q,T,A)$ and the Solver takes $(Q,A)$. The learnable vector is injected into the Solver representation as
\begin{equation}
\widetilde{\boldsymbol{x}}_l
=
\boldsymbol{x}_l
+
\boldsymbol{\theta}_l.
\label{eq:learnable_vector_injection}
\end{equation}
We do not use a separate scaling coefficient in learnable adaptation, because the magnitude of $\boldsymbol{\theta}_l$ is learned during optimization.

Let \(z_l^R\) denote the support-set averaged answer-token representation of the Reasoner path at layer \(l\). Similarly, let \(z_l^S\) denote the support-set averaged answer-token representation of the Solver path after vector injection. The learnable vector is optimized to bring the intervened Solver representation closer to the Reasoner representation:
\begin{equation}
\mathcal{L}_{\mathrm{rec}}
=
% \sum_{l\in\mathcal{L}_{\mathrm{train}}}
\left\|
\boldsymbol{z}_l^{\mathcal{R}}
-
\boldsymbol{z}_l^{\mathcal{S}}
\right\|_2^2,
\qquad
\mathcal{L}=\lambda_{\mathrm{rec}}\mathcal{L}_{\mathrm{rec}}.
\label{eq:rec_loss}
\end{equation}

Here, \(\lambda_{\mathrm{rec}}\) controls the weight of the representation matching loss. During adaptation, the VLM backbone remains frozen, and only the selected-layer learnable vector parameter \(\theta_l\) is updated.

\subsection{Layer Selection and Injection}
\label{subsec:injection_layer_selection}

Because activation intervention is layer-sensitive~\citep{todd2023function}, we select the injection layer using a small held-out subset before final evaluation. Specifically, for each benchmark, we inject the Reasoning Vector into each candidate language-model layer separately and evaluate performance on this subset. The layer with the best held-out performance is then used for full evaluation. We apply the same layer-selection protocol to all vector sources and adaptation settings.

At inference time, LIFT applies vector addition during autoregressive decoding at the selected language-model layer. Let $x_l$ denote the hidden state at the selected layer. Static and learnable interventions are given by:
\begin{equation}
\tilde{x}_l = x_l + \mu r_l, 
\qquad
\tilde{x}_l = x_l + \theta_l,
\end{equation}
where $\mu$ controls the strength of the static Reasoning Vector and $\theta_l$ is the optimized learnable vector. LIFT only modifies language-token hidden states, without updating VLM parameters.

\section{Experiments}
\label{sec:experiments}

We evaluate LIFT on Qwen2.5-VL-7B-Instruct~\citep{bai2025qwen25vltechnicalreport} and InternVL2.5-8B~\citep{chen2024expanding}, which are built on Qwen2.5 and InternLM2.5 language backbones, across six reasoning benchmarks. Our experiments study how language-side Reasoning Vectors behave after multimodal scaling. We first test whether static vector intervention can improve VLM reasoning, and then compare Reasoning Vectors extracted from different sources: LLM-derived, VLM-derived, and VLM-MM. Finally, we study learnable vector adaptation to examine whether the gains come from the semantic structure of the extracted Reasoning Vectors, rather than simply from introducing additional trainable parameters. We describe the experimental setup in Section~\ref{subsec:experimental_setup} and present the main results in Section~\ref{subsec:main_results}, followed by representation-level and layer-wise analyses in Section~\ref{subsec:representation_layerwise_analysis}.

\subsection{Setup and Implementation Details}
\label{subsec:experimental_setup}

\noindent\textbf{Models and datasets.}
We evaluate LIFT on two VLMs across six reasoning benchmarks that span text-only and multimodal settings. The text-only benchmarks are GSM8K~\citep{cobbe2021gsm8k}, CommonsenseQA~\citep{talmor2019commonsenseqa}, and StrategyQA~\citep{geva2021did}, while the multimodal benchmarks are MathVista~\citep{lu2023mathvista}, MathVision~\citep{wang2024mathvision}, and ScienceQA~\citep{lu2022scienceqa}. This benchmark suite allows us to examine whether language-side Reasoning Vectors remain effective not only in pure-text reasoning, but also in multimodal reasoning tasks with VLMs. More details on the datasets and evaluation protocols are provided in Appendix~\ref{app:datasets_evaluation}.
% \vspace{-1.0em}

\noindent\textbf{Implementation details.}
For a fair comparison, all LIFT variants use the same zero-shot CoT inference setup as the baseline VLM; they differ only by the injected Reasoning Vector and the selected intervention layer. Vector extraction follows the source definitions in Section ~\ref{subsec:vector_sources}. For each evaluation benchmark, we extract the Reasoning Vector from a corresponding support set: GSM8K, MathVista, and MathVision use a GSM8K support set; CommonsenseQA and StrategyQA use support sets from their own training splits; and ScienceQA uses a CommonsenseQA support set.

For static vector transfer, the intervention strength $\mu$ is fixed to $1.0$. For learnable vector adaptation, only the injected vectors are optimized, and the learning rate is set to $1\times10^{-4}$. For the layer-selection procedure described in Section~\ref{subsec:injection_layer_selection}, we randomly sample 100 training examples as a development subset to choose the best injection layer, and then use the selected layer for full evaluation.

\subsection{Main Results}
\label{subsec:main_results}

\subsubsection{Reasoning Vectors Transfer Effectively Across Modalities}
\label{subsubsec:overall_effectiveness}

\begin{table}[t]
\centering
\small
\setlength{\tabcolsep}{3.8pt}
\renewcommand{\arraystretch}{1.12}
\caption{
Main results on six reasoning benchmarks across two VLMs. We abbreviate MathVista/MathVision/CommonsenseQA/StrategyQA/ScienceQA as MVista/MVision/CSQA/SQA/SciQA. LLM/VLM-derived denote static vector transfer using Reasoning Vectors extracted from the base LLM and the corresponding VLM, respectively; LLM/VLM-learnable are learnable vectors initialized from these corresponding static vectors. Bold and underlined numbers indicate the best and second-best results for each model and benchmark. All results are reported as accuracy (\%), and Avg. denotes the average accuracy across benchmarks.  
}

\label{tab:main_results}
\vspace{0.3em}
\resizebox{0.9\textwidth}{!}{%
\begin{tabular*}{\textwidth}{@{\extracolsep{\fill}}llccccccc@{}}
\toprule[1.2pt]
\textbf{Model} & \textbf{Method}
& \textbf{GSM8K} & \textbf{MVista} & \textbf{MVision}
& \textbf{CSQA} & \textbf{SQA} & \textbf{SciQA} & \textbf{Avg.} \\
\midrule

\multirow{8}{*}{Qwen2.5-VL}
& Baseline    & 82.2 & 68.5 & 25.0 & 77.3 & 58.6 & 87.8 & 66.6 \\
\cmidrule(l){2-9}
& 1-shot CoT  & 82.7 & 68.4 & 25.3 & 75.1 & 61.2 & 87.9 & 66.8 \\
& 2-shot CoT  & 82.9 & 68.4 & 25.3 & 76.8 & 61.3 & 87.9 & 67.1 \\
& 4-shot CoT  & 82.9 & 68.4 & 25.3 & \underline{78.1} & 61.6 & 87.9 & 67.4 \\
\cmidrule(l){2-9}
& LLM-derived    & \underline{84.2} & 69.1 & \underline{26.6} & 78.0 & \textbf{63.5} & \underline{88.1} & \underline{68.3} \\
& VLM-derived    & 83.0 & \underline{70.7} & 25.7 & 77.7 & 60.3 & 87.8 & 67.5 \\
\cmidrule(l){2-9}
& LLM-learnable  & \textbf{85.6} & \textbf{71.8} & \textbf{27.3} & \textbf{78.3} & \underline{62.4} & \textbf{88.9} & \textbf{69.1} \\
& VLM-learnable  & 83.2 & 69.5 & 24.7 & 77.7 & 60.3 & 87.9 & 67.2 \\

\midrule[1.2pt]

\multirow{8}{*}{InternVL2.5}
& Baseline    & 77.1 & 62.8 & 22.7 & 82.3 & 65.6 & 96.5 & 67.8 \\
\cmidrule(l){2-9}
& 1-shot CoT  & 77.1 & 60.3 & 23.4 & 82.4 & 65.7 & 96.7 & 67.6 \\
& 2-shot CoT  & 77.2 & 63.8 & 23.4 & 82.7 & 63.6 & 96.7 & 67.9 \\
& 4-shot CoT  & 77.4 & 63.8 & 23.4 & \underline{83.1} & 66.7 & 96.7 & 68.5 \\
\cmidrule(l){2-9}
& LLM-derived    & \underline{78.1} & \underline{64.0} & \underline{25.0} & \underline{83.1} & \underline{67.1} & \underline{96.8} & \underline{69.0} \\
& VLM-derived    & 77.9 & 63.0 & 23.4 & 82.6 & 65.9 & 96.7 & 68.3 \\
\cmidrule(l){2-9}
& LLM-learnable  & 78.0 & \textbf{64.6} & \textbf{25.3} & \textbf{83.2} & \textbf{67.6} & \textbf{97.4} & \textbf{69.4} \\
& VLM-learnable  & \textbf{78.4} & 63.4 & 20.1 & 79.0 & 62.4 & 96.0 & 66.6 \\

\bottomrule[1.2pt]
\end{tabular*}
}
\vspace{-1.2em}
\end{table}

Both LLM-derived and VLM-derived vectors improve the average score over the baseline, with LLM-derived vectors giving stronger and more consistent gains. As shown in Table~\ref{tab:main_results}, LLM-derived vectors improve the average score from 66.6/67.8 to 68.3/69.0 on Qwen2.5-VL/InternVL2.5. Notably, LLM-derived vectors outperform 4-shot CoT in average score on both VLMs, while keeping the zero-shot CoT inference setup unchanged except for vector injection. These results show that static Reasoning Vector intervention can improve VLM reasoning without updating model parameters. Appendix~\ref{app:bootstrap_ci} provides split sizes and bootstrap confidence intervals for the main accuracy gains. Appendix~\ref{app:same_norm_random} uses same-norm random vectors to test whether the extracted direction matters, beyond simply adding a vector with the same norm.

We further observe cross-modal transfer. The Reasoning Vectors evaluated on multimodal benchmarks in Table~\ref{tab:main_results} are extracted from text-only support sets. Specifically, MathVista and MathVision use vectors extracted from GSM8K, while ScienceQA uses vectors extracted from CommonsenseQA. Despite this text-only extraction, LLM-derived vectors improve all three multimodal benchmarks on both VLMs. On MathVista/MathVision, they raise scores from 68.5/25.0 to 69.1/26.6 on Qwen2.5-VL-7B-Instruct and from 62.8/22.7 to 64.0/25.0 on InternVL2.5-8B. These results indicate that language-side reasoning vectors extracted from text-only data can still improve multimodal reasoning performance. 

\subsubsection{LLM-derived Vectors Provide More Reliable Interventions than VLM-derived Vectors}
\label{subsubsec:vector_source}

Vector source clearly affects performance. LLM-derived vectors outperform VLM-derived vectors on both VLMs, reaching 68.3/69.0 versus 67.5/68.3 on Qwen2.5-VL/InternVL2.5, and achieve higher scores on most datasets. This suggests that VLM-derived vectors are still useful but less reliable under our extraction protocol. A possible explanation is that multimodal alignment changes the organization of language-side reasoning representations in the VLM, making VLM-extracted directions less stable as interventions than reasoning-conditioned directions extracted from the base LLM.

\begin{figure}[t]
    \centering
    \includegraphics[width=0.9\linewidth]{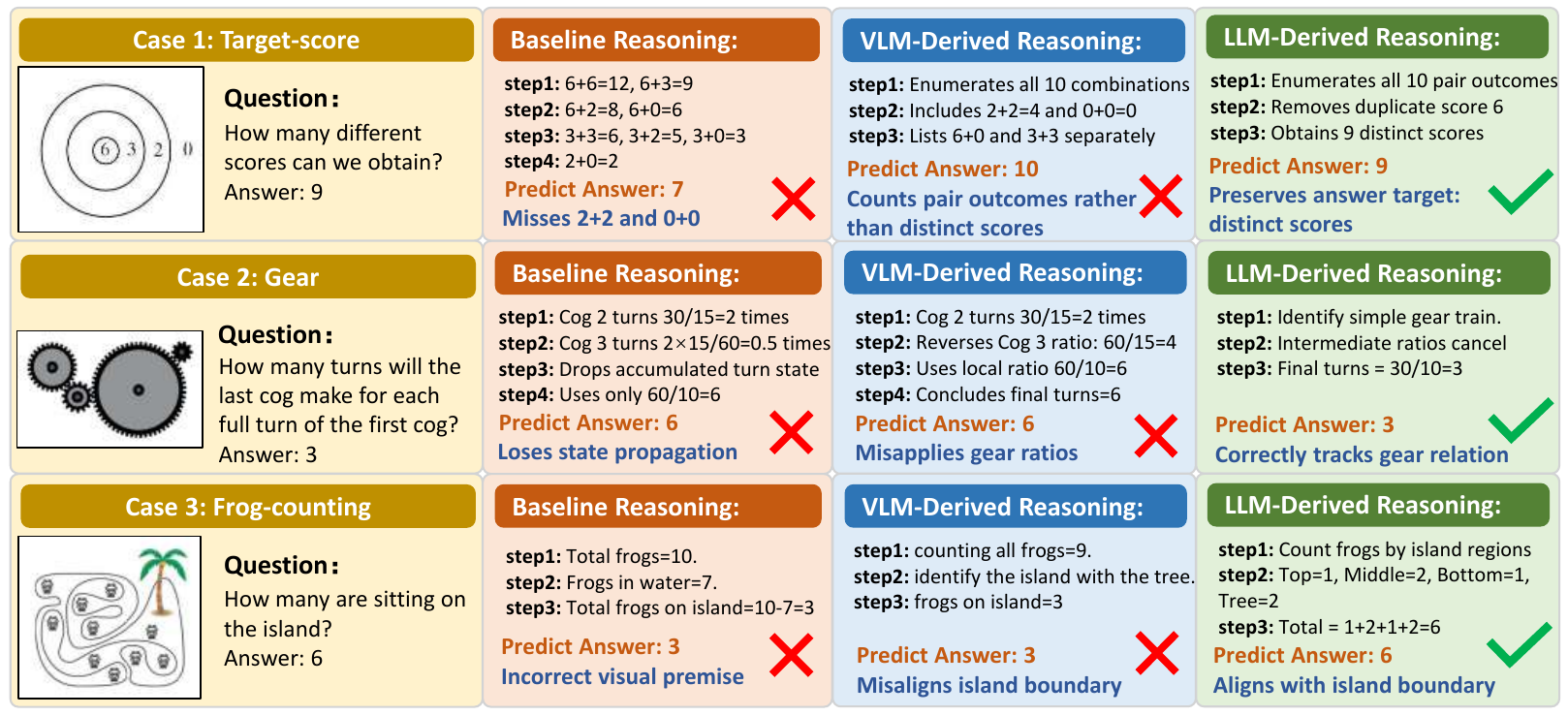}
    \caption{Representative Case Studies of Reasoning Repairs.}
    \vspace{-1.0em}
    \label{fig:case}
\end{figure}

We further examine representative MathVision cases to understand how vector source affects multimodal reasoning. As shown in Figure~\ref{fig:case}, LLM-derived vectors often repair intermediate reasoning errors rather than merely changing the final answer. In \emph{target-score}, the baseline misses repeated-hit cases such as \(2+2\) and \(0+0\), while the VLM-derived vector counts pair outcomes instead of distinct scores. The LLM-derived vector preserves both repeated-hit enumeration and score deduplication, giving the correct answer 9. In \emph{gear}, the baseline and VLM-derived vector use local gear ratios but fail to propagate the full chain, whereas the LLM-derived vector preserves the end-to-end relation. In \emph{frog-counting}, the baseline relies on a subtractive count from frogs assumed to be in the water, while the VLM-derived vector narrows the island to the palm-tree region. The LLM-derived vector instead follows the island boundary and sums frogs by region. These cases suggest that LLM-derived vectors can influence intermediate reasoning behavior and help keep the response aligned with the question target. Additional examples with full reasoning traces are provided in Appendix~\ref{app:case_study}.

Overall, the quantitative results and case studies suggest that LLM-derived vectors provide \textbf{more reliable interventions} than VLM-derived vectors. Rather than merely changing final predictions, they appear to help preserve reasoning steps that remain aligned with the target question.

\subsubsection{LLM-derived Vectors Outperform VLM-MM Vectors}
\label{subsubsec:multimodal_extraction}

% Table~\ref{tab:vlm_multimodal_source} compares VLM-side extraction with text-only and multimodal support sets. Compared with VLM-derived vectors, VLM-MM improves the average score from 56.5/59.4 to 57.0/60.6 on Qwen2.5-VL/InternVL2.5. The gain mainly comes from MathVision, where scores increase from 25.3/22.1 to 26.0/24.7. This suggests that, with the source model fixed to the VLM, vectors derived from the same multimodal benchmark outperform those extracted from text-only support sets.

% However, LLM-derived vectors still achieve the best average performance. Across MathVision and ScienceQA, the averages are 57.4/60.9 for LLM-derived, 57.0/60.6 for VLM-MM, and 56.5/59.4 for VLM-derived on Qwen2.5-VL/InternVL2.5. This comparison gives VLM-side extraction a favorable setting, since VLM-MM extracts vectors from the target VLM using support sets from the same multimodal benchmark used for evaluation. The lower performance suggests that the advantage of LLM-derived vectors comes from the source model, not just the extraction data.

Table~\ref{tab:vlm_multimodal_source} compares VLM-side extraction with text-only and multimodal support sets. Compared with VLM-derived vectors, VLM-MM improves the average score from 56.8/60.1 to 57.0/60.6 on Qwen2.5-VL/InternVL2.5. The gain mainly comes from MathVision, where scores increase from 25.7/23.4 to 26.0/24.7. This suggests that, with the source model fixed to the VLM, vectors derived from the same multimodal benchmark are more effective on average than those extracted from text-only support sets.

However, LLM-derived vectors still achieve the best average performance. Across MathVision and ScienceQA, the averages are 57.4/60.9 for LLM-derived, 57.0/60.6 for VLM-MM, and 56.8/60.1 for VLM-derived on Qwen2.5-VL/InternVL2.5. This comparison gives VLM-side extraction a favorable setting, since VLM-MM extracts vectors from the target VLM using support sets from the same multimodal benchmark used for evaluation. The lower performance suggests that the advantage of LLM-derived vectors comes from the source model, not just the extraction data.

\begin{wraptable}{r}{0.5\textwidth}
\vspace{-1.5em}
\centering
\scriptsize
\setlength{\tabcolsep}{2.6pt}
\renewcommand{\arraystretch}{1.06}
\caption{
Accuracy (\%) of VLM multimodal extraction on multimodal reasoning benchmarks.
}
\label{tab:vlm_multimodal_source}
\vspace{0.1em}
\begin{tabular}{llccc}
\toprule
\textbf{Model} & \textbf{Method} & \textbf{MVision} & \textbf{SciQA} & \textbf{Avg.} \\
\midrule
\multirow{4}{*}{Qwen2.5-VL}
& Baseline    & 25.0 & 87.8 & 56.4 \\
& LLM-derived & \textbf{26.6} & \textbf{88.1} & \textbf{57.4} \\
& VLM-derived & 25.7 & 87.8 & 56.8 \\
& VLM-MM      & \underline{26.0} & \underline{87.9} & \underline{57.0} \\
\midrule
\multirow{4}{*}{InternVL2.5}
& Baseline    & 22.7 & 96.5 & 59.6 \\
& LLM-derived & \textbf{25.0} & \textbf{96.8} & \textbf{60.9} \\
& VLM-derived & 23.4 & \underline{96.7} & 60.1 \\
& VLM-MM      & \underline{24.7} & 96.4 & \underline{60.6} \\
\bottomrule
\end{tabular}
\vspace{-1.5em}
\end{wraptable}

This ordering, LLM-derived $>$ VLM-MM $>$ VLM-derived, suggests both extraction data and source model matter. VLM-MM is more effective than VLM-derived on the corresponding multimodal benchmarks, but still remains below LLM-derived on average. One possible explanation is that multimodal alignment changes the organization of language-side reasoning representations in the VLM. As a result, even language-side vectors extracted from the VLM may already reflect multimodal alignment effects, making VLM-MM useful for the corresponding multimodal benchmark but less stable as a general steering direction than pre-alignment LLM-derived vectors.

\vspace{-1.0em}
\subsubsection{Meaningful Initialization Matters for Learnable Vector Adaptation}
\label{subsubsec:learnable_adaptation}

Learnable vector adaptation is most effective when initialized from LLM-derived vectors. As shown in Table~\ref{tab:main_results}, LLM-learnable achieves the best average score on both VLMs, improving static LLM-derived vectors from 68.3/69.0 to 69.1/69.4 on Qwen2.5-VL/InternVL2.5. This suggests that LLM-derived vectors already provide useful reasoning directions, and lightweight training can further adapt them to the target VLM.

The comparison between LLM-learnable and VLM-learnable shows that the initialization source matters. Although both settings optimize injected vectors with the VLM backbone fixed, LLM-learnable consistently outperforms VLM-learnable, reaching 69.1/69.4 versus 67.2/66.6 on Qwen2.5-VL/InternVL2.5. On InternVL2.5-8B, VLM-learnable even falls below the baseline, indicating that learnable adaptation is sensitive to initialization quality.

\begin{wraptable}{r}{0.60\textwidth}
\vspace{-1.0em}
\centering
\tiny
\setlength{\tabcolsep}{2.0pt}
\renewcommand{\arraystretch}{1.05}
\caption{
Initialization ablation for learnable vector adaptation. Random init replaces the extracted Reasoning Vector with a random vector under the same training setting.
}
\label{tab:init_ablation}
\vspace{0.3em}
\resizebox{\linewidth}{!}{
\begin{tabular}{llccccccc}
\toprule
\textbf{Model} & \textbf{Method}
& \textbf{GSM8K} & \textbf{MVista} & \textbf{MVision}
& \textbf{CSQA} & \textbf{SQA} & \textbf{SciQA} & \textbf{Avg.} \\
\midrule

\multirow{4}{*}{Qwen2.5-VL}
& LLM-learnable & \textbf{85.6} & \textbf{71.8} & \textbf{27.3} & \textbf{78.3} & \textbf{62.4} & \textbf{88.9} & \textbf{69.1} \\
& Random init   & 83.2 & 68.6 & 24.7 & 77.4 & 58.8 & 87.9 & 66.7 \\
\cmidrule(l){2-9}
& VLM-learnable & \textbf{83.2} & \textbf{69.5} & \textbf{24.7} & \textbf{77.7} & \textbf{60.3} & \textbf{87.9} & \textbf{67.2} \\
& Random init   & 83.0 & 68.6 & \textbf{24.7} & 77.4 & 58.8 & \textbf{87.9} & 66.7 \\

\midrule

\multirow{4}{*}{InternVL2.5}
& LLM-learnable & \textbf{78.0} & \textbf{64.6} & \textbf{25.3} & \textbf{83.2} & \textbf{67.6} & \textbf{97.4} & \textbf{69.4} \\
& Random init   & 77.4 & 63.3 & 24.1 & 83.1 & 65.4 & 96.7 & 68.3 \\
\cmidrule(l){2-9}
& VLM-learnable & 78.4 & \textbf{63.4} & \textbf{20.1} & 79.0 & \textbf{62.4} & \textbf{96.0} & \textbf{66.6} \\
& Random init   & \textbf{78.8} & 62.5 & 19.7 & \textbf{79.4} & 59.7 & 95.8 & 66.0 \\

\bottomrule
\end{tabular}
}
\vspace{-1.0em}
\end{wraptable}

Table~\ref{tab:init_ablation} further compares LLM-derived initialization with random initialization under the same adaptation protocol. Replacing LLM-derived initialization with random initialization reduces the average score from 69.1/69.4 to 66.7/68.3 on Qwen2.5-VL/InternVL2.5. This shows that the gains of LLM-learnable are not merely due to adding trainable activation-level parameters, but also rely on a useful language-side reasoning-direction initialization.

A possible explanation is that VLM-derived vectors are affected by multimodal alignment even when computed from language-side answer states. They may therefore be less stable as initialization. In contrast, LLM-derived vectors provide a cleaner language-side starting point, so adaptation mainly aligns an existing reasoning direction to the target VLM rather than discovering one from scratch.

\subsection{Representation Alignment and Layer-wise Analysis}
\label{subsec:representation_layerwise_analysis}

\begin{figure}[!t]
    \centering
    \includegraphics[width=0.8\linewidth]{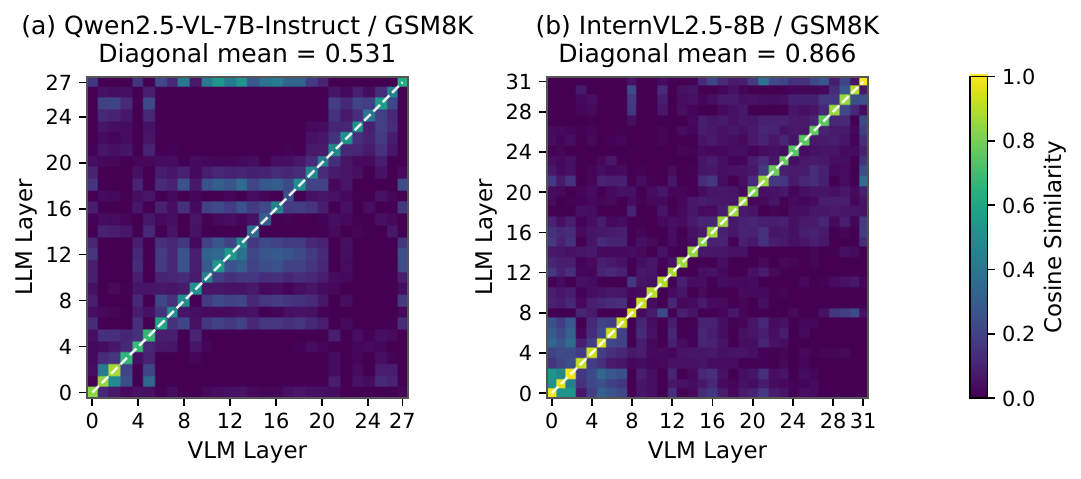}
    \caption{
    Cross-layer cosine similarity between LLM- and VLM-derived vectors on GSM8K.
    }
    \label{fig:cos}
\end{figure}

\noindent\textbf{Cross-layer vector alignment.}
The cosine similarity between LLM-derived and VLM-derived vectors in Fig.~\ref{fig:cos} shows diagonal concentration in both heatmaps, indicating that vectors from similar depths align better between the base LLM and the corresponding VLM. This suggests that multimodal scaling partially preserves the layer-wise organization of language-side reasoning representations.

However, the strength of this correspondence differs across models. InternVL2.5-8B exhibits a strong diagonal pattern on GSM8K, with an average diagonal similarity of 0.866, suggesting close layer-wise LLM--VLM alignment. In contrast, Qwen2.5-VL-7B-Instruct shows a more diffuse pattern, with an average diagonal similarity of 0.531 and stronger off-diagonal similarities in middle layers. This suggests that reasoning-related structures are partially preserved but may be redistributed or less cleanly aligned after multimodal scaling. Overall, this layer-wise correspondence helps explain why LLM-derived Reasoning Vectors can transfer to VLMs, while the weaker and more diffuse patterns also motivate layer-specific intervention and layer selection.

\begin{figure}[!t]
    \centering
    \includegraphics[width=0.90\linewidth]{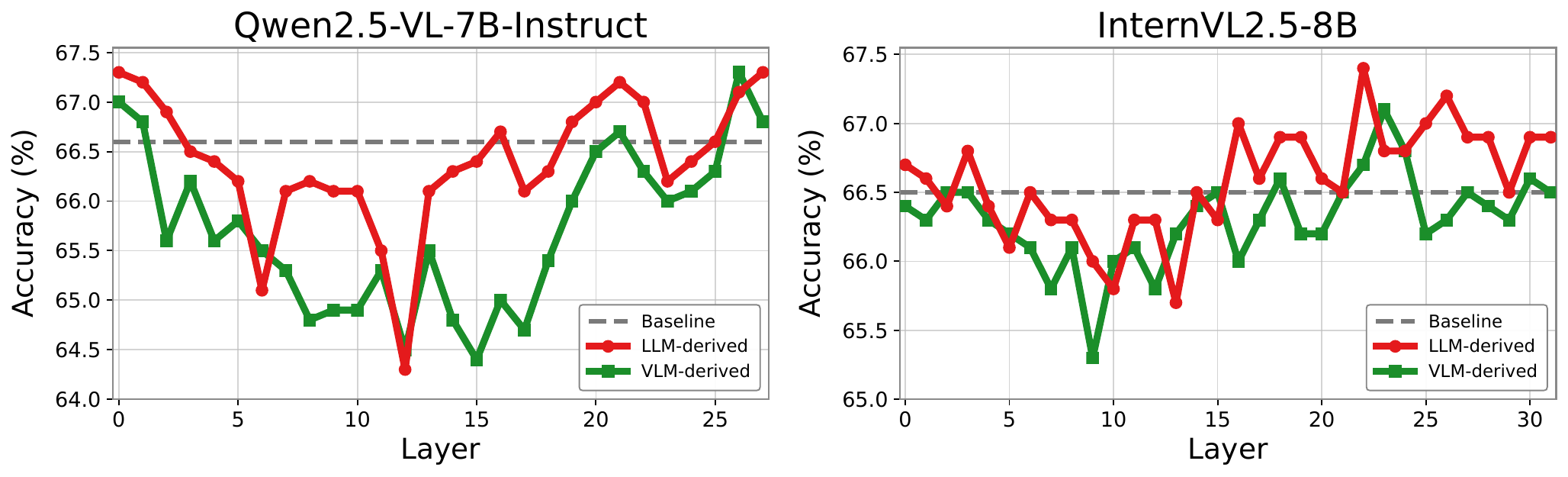}
    \caption{
    Layer-wise performance of LLM- and VLM-derived vectors averaged over six datasets.
    }
    \vspace{-0.9em}
    \label{fig:layerwise}
\end{figure}

\noindent\textbf{Layer-wise sensitivity.}
Fig.~\ref{fig:layerwise} shows Reasoning Vector injection is strongly layer-sensitive. Both LLM-derived and VLM-derived vectors fluctuate across layers, suggesting their effect depends on layer-specific VLM organization rather than applying uniformly at all depths. Despite this variation, the source-level trend remains clear: LLM-derived vectors generally outperform VLM-derived vectors across layers. This is consistent with Table~\ref{tab:main_results}, where VLM-derived vectors remain useful but LLM-derived vectors provide stronger and more reliable interventions after multimodal scaling.

\section{Conclusion and Limitations}
\label{sec:conclusion}

We presented LIFT, a lightweight vector-intervention method that transfers language-side Reasoning Vectors from a base LLM to its corresponding VLM. Across text-only and multimodal reasoning benchmarks, LIFT improves VLM reasoning without updating the VLM backbone. These results suggest that reasoning-related directions formed in the base LLM can remain usable after multimodal scaling, even though multimodal alignment may partially weaken or reorganize them in the corresponding VLM.

LIFT also has several limitations. Our experiments cover two VLM families, and future work should evaluate whether the findings generalize to more architectures, alignment procedures, and multimodal tasks. LIFT requires support examples to extract Reasoning Vectors, and its effectiveness is sensitive to the selected intervention layer. In addition, because LIFT operates on language-side hidden states, it cannot directly correct errors caused by inaccurate visual perception or grounding. A promising direction is to combine reasoning-vector intervention with stronger visual grounding methods.

\section*{Acknowledgements}

This work is supported by the National Natural Science Foundation
of China (62576091), Jiangsu Province Carbon Peak Carbon Neutrality
Science and Technology Innovation Special Fund Project
(Grant No. BT2025029), the Southeast University Big Data Computing
Center, and the Southeast University Kunpeng \& Ascend Center
of Cultivation. The authors declare no competing interests.

\clearpage
\bibliographystyle{plainnat}
\bibliography{references}

\clearpage
\appendix
\appendix

\section{Derivation of Reasoning Vector}
\label{app:derivation}

When the model reasons with a reasoning trace, the generation of an answer token $a \in A$ depends not only on the question tokens but also on the intermediate reasoning tokens.
Let $Q$, $T$, and $A$ denote the question, reasoning trace, and answer tokens, respectively.
For a simplified single-head self-attention module, we denote the corresponding key and value matrices as $(K_Q,V_Q)$, $(K_T,V_T)$, and $(K_A,V_A)$.
The self-attention output for answer token $a$ under the reasoning-conditioned input $(Q,T,A)$ can be written as
\begin{align}
&\mathrm{SA}\left(a,[K_Q,K_T,K_A],[V_Q,V_T,V_A]\right) \nonumber \\
&=
\frac{\exp(aK_Q^{\top})}{Z_{\mathrm{total}}}V_Q
+
\frac{\exp(aK_T^{\top})}{Z_{\mathrm{total}}}V_T
+
\frac{\exp(aK_A^{\top})}{Z_{\mathrm{total}}}V_A \nonumber \\
&=
\frac{Z_Q}{Z_{\mathrm{total}}}
\cdot
\underbrace{
\frac{\exp(aK_Q^{\top})}{Z_Q}V_Q
}_{\mathrm{SA}(a,[K_Q],[V_Q])}
+
\frac{Z_T}{Z_{\mathrm{total}}}
\cdot
\underbrace{
\frac{\exp(aK_T^{\top})}{Z_T}V_T
}_{\mathrm{SA}(a,[K_T],[V_T])}
+
\frac{Z_A}{Z_{\mathrm{total}}}
\cdot
\underbrace{
\frac{\exp(aK_A^{\top})}{Z_A}V_A
}_{\mathrm{SA}(a,[K_A],[V_A])}.
\label{eq:app_attention_with_reasoning}
\end{align}

where \( Z_Q = \sum \exp(aK_Q^{\top}),\; Z_T = \sum \exp(aK_T^{\top}),\; Z_A = \sum \exp(aK_A^{\top}), \) and \( Z_{\mathrm{total}} = Z_Q + Z_T + Z_A. \)

In contrast, without the reasoning trace, the answer token only attends to the question and answer tokens.
The self-attention output under the input $(Q,A)$ is
\begin{equation}
\mathrm{SA}\left(a,[K_Q,K_A],[V_Q,V_A]\right)
=
\frac{Z_Q}{Z_Q+Z_A}
\cdot
\mathrm{SA}(a,[K_Q],[V_Q])
+
\frac{Z_A}{Z_Q+Z_A}
\cdot
\mathrm{SA}(a,[K_A],[V_A]).
\label{eq:app_attention_without_reasoning}
\end{equation}

By comparing Eq.~\ref{eq:app_attention_with_reasoning} and Eq.~\ref{eq:app_attention_without_reasoning}, the reasoning-conditioned attention output can be written as a standard attention term plus an additive reasoning-conditioned shift:
\begin{align}
&\mathrm{SA}\left(a,[K_Q,K_T,K_A],[V_Q,V_T,V_A]\right) \nonumber \\
&=
\underbrace{
\mathrm{SA}\left(a,[K_Q,K_A],[V_Q,V_A]\right)
}_{\text{Standard Attention}}
+
\mu \cdot
\underbrace{
\left(
\mathrm{SA}(a,[K_T],[V_T])
-
\mathrm{SA}\left(a,[K_Q,K_A],[V_Q,V_A]\right)
\right)
}_{\text{Reasoning Vector}},
\label{eq:app_reasoning_shift_attention}
\end{align}
where $\mu = \frac{Z_T}{Z_{\mathrm{total}}}$.

Equation~\ref{eq:app_reasoning_shift_attention} shows that, in this simplified setting, the reasoning trace acts as an additive term in the attention output of the answer tokens. In other words, the contribution of the reasoning to answer generation can be summarized as an additive change in the model's internal states. LIFT operationalizes this view through answer-token hidden-state differences under the Reasoner--Solver formulation. Recent theoretical work \cite{liang2026understandingoutofdistributiongeneralizationchainofthought} characterizes Chain-of-Thought reasoning as compositional out-of-distribution generalization, where reusable single-step reasoning transitions are composed to solve longer tasks. This perspective is complementary to our derivation: while prior work studies the generalization mechanism of multi-step reasoning, our analysis focuses on the representation-level changes induced by explicit reasoning traces, motivating the construction of Reasoning Vectors through the Reasoner--Solver difference.

\section{Implementation Details}
\label{app:implementation_details}

\subsection{Datasets and Evaluation Protocols}
\label{app:datasets_evaluation}

We use Qwen2.5-7B-Instruct as the base LLM for Qwen2.5-VL-7B-Instruct, and InternLM2.5-7B-Chat as the base LLM for InternVL2.5-8B. We evaluate LIFT on six reasoning benchmarks: GSM8K, MathVista, MathVision, CommonsenseQA, StrategyQA, and ScienceQA. These benchmarks cover both text-only and multimodal reasoning scenarios. For each benchmark, we use a held-out evaluation split for final testing. Unless otherwise specified, the main evaluation follows the zero-shot CoT setting used in the main table. All methods are evaluated under the same zero-shot CoT prompt, decoding setup, evaluation split, and answer parser. Support sets are used only before evaluation to extract Reasoning Vectors or adapt learnable vectors, and are never included in the test set.

\paragraph{GSM8K.}
GSM8K is a text-only mathematical reasoning benchmark. We use the training split to construct support examples and evaluate on the test split. The final numerical answer is parsed from the model response, with \verb|\boxed{}| used as the preferred answer format whenever applicable.

\paragraph{MathVista.}
MathVista is a multimodal mathematical reasoning benchmark. We evaluate on the testmini split, which contains 1,000 examples.

\paragraph{MathVision.}
MathVision is a visually grounded mathematical reasoning benchmark. It contains problems that require understanding diagrams, charts, geometric structures, or other visual contexts before performing mathematical reasoning. We follow the evaluation split and answer parsing protocol used in the main experiments, and parse the final answer from the model response.

\paragraph{CommonsenseQA.}
CommonsenseQA is a text-only multiple-choice commonsense reasoning benchmark. We evaluate on 1,221 examples.

\paragraph{StrategyQA.}
StrategyQA is a text-only binary reasoning benchmark that requires implicit multi-hop reasoning. The model is required to produce a final answer in a constrained yes-or-no format. Accuracy is computed by matching the parsed answer with the ground-truth label.

\paragraph{ScienceQA.}
ScienceQA is a multimodal science reasoning benchmark. It contains questions that may require scientific knowledge, visual understanding, or both. We evaluate on the subset used in the main experiments and follow the same answer parsing protocol.

\paragraph{Scope of comparisons.}
LIFT is designed as a lightweight intervention and analysis tool for studying whether language-side Reasoning Vectors from a base LLM remain usable inside the corresponding VLM after multimodal scaling. Our main comparison therefore focuses on different vector sources under the same protocol. We do not aim to replace training-based VLM adaptation methods or reinforcement learning. 

The average score reported in the main tables is computed as the arithmetic mean over the evaluated benchmarks. Table~\ref{tab:app_dataset_support} summarizes the evaluation sizes and support sources used for each benchmark.

\begin{table}[htbp]
\centering
\footnotesize
\caption{
Evaluation split sizes and support sources used in our experiments. Support examples are used only for offline Reasoning Vector extraction or learnable adaptation and are never appended to evaluation prompts.
}
\label{tab:app_dataset_support}
\begin{tabular}{lccc}
\toprule
Benchmark & Modality & \#Eval. examples & Support source \\
\midrule
GSM8K & Text & 1319 & GSM8K \\
MathVista & Image-text & 1000 & GSM8K \\
MathVision & Image-text & 304 & GSM8K \\
CommonsenseQA & Text & 1221 & CommonsenseQA \\
StrategyQA & Text & 687 & StrategyQA \\
ScienceQA & Image-text & 4241 & CommonsenseQA \\
\bottomrule
\end{tabular}
\end{table}

\FloatBarrier

\subsection{Prompting and Generation Settings}
\label{app:prompt_generation}

During evaluation, the baseline and all LIFT variants use the same zero-shot CoT setting, decoding configuration, and answer parser. Support examples and extracted traces are used only before evaluation for vector extraction or learnable adaptation; they are not included in the evaluation set or used as in-context demonstrations. Therefore, the reported performance differences are not caused by prompt changes.

For Reasoning Vector extraction, we construct two teacher-forced paths from each support set. The Reasoner path receives the question, a model-generated reasoning trace, and the final answer, while the Solver path receives only the question and the final answer. This with-CoT versus without-CoT comparison is used only offline to compute the answer-token hidden-state difference that defines the Reasoning Vector.

For mathematical tasks, we ask the model to place the final answer within \verb|\boxed{}| whenever applicable. For multiple-choice tasks, we parse the predicted option. For yes-or-no tasks, we parse the final yes-or-no answer. For all LIFT evaluations, we use the same evaluation prompt as the corresponding non-intervened baseline, so that performance changes can be attributed to the internal intervention rather than to prompt changes. Table~\ref{tab:app_prompt_templates} lists the prompt templates used in our experiments.

\begin{table}[htbp]
\centering
\small
\setlength{\tabcolsep}{4pt}
\renewcommand{\arraystretch}{0.92}
\caption{Prompt templates used in our experiments.}
\label{tab:app_prompt_templates}
\begin{tabular}{@{}p{0.17\linewidth} p{0.66\linewidth}@{}}
\toprule
Setting & Prompt Template \\
\midrule
Direct Answer &
\texttt{Please answer the question directly. Put your final answer within \textbackslash boxed\{\}.} \\
\midrule
Reasoning Trace &
\texttt{Please reason step by step. Put your final answer within \textbackslash boxed\{\}.} \\
\midrule
Multiple Choice &
\texttt{Please choose the correct option. Put your final answer within \textbackslash boxed\{\}.} \\
\midrule
Yes/No &
\texttt{Please answer the question with yes or no. Put your final answer within \textbackslash boxed\{\}.} \\
\bottomrule
\end{tabular}
\end{table}

For all baseline and LIFT evaluations, we use deterministic decoding with beam size 1 and sampling disabled. The maximum number of new tokens is set to 512 for text-only benchmarks and 4096 for multimodal benchmarks, as summarized in Table~\ref{tab:app_generation_config}.

\begin{table}[htbp]
\centering
\caption{Generation configuration used for evaluation.}
\label{tab:app_generation_config}
\begin{tabular}{lc}
\toprule
Parameter & Value \\
\midrule
Number of beams & 1 \\
Maximum new tokens & 512 / 4096 \\
Length penalty & 0.0 \\
Random seed & 42 \\
\bottomrule
\end{tabular}
\end{table}

\FloatBarrier

\subsection{Support Set Construction and Vector Extraction}
\label{app:support_vector_extraction}
Unless otherwise specified, we sample 500 support examples for each support source. Each support example contains a question $Q$, a model-generated reasoning trace $T$, and a final answer $A$. The reasoning trace $T$ is generated by the corresponding source model under the step-by-step CoT prompt.

For each vector source, we keep only support examples that are answered correctly by that source model. Thus, LLM-derived vectors are extracted from examples correctly answered by the corresponding base LLM, while VLM-derived vectors are extracted from examples correctly answered by the target VLM. As a result, LLM-derived and VLM-derived vectors may be extracted from different correct-example subsets, but they follow the same Reasoner--Solver extraction protocol.

For GSM8K, MathVista, and MathVision, we use GSM8K support examples to extract mathematical reasoning vectors. For CommonsenseQA and StrategyQA, we use support examples from their corresponding training splits. For ScienceQA, we use CommonsenseQA support examples to evaluate whether language-side commonsense reasoning vectors transfer to multimodal science reasoning. These support-source choices are summarized together with the evaluation sizes in Table~\ref{tab:app_dataset_support}.
\FloatBarrier

\subsection{Learnable Vector Adaptation}
\label{app:learnable_adaptation_details}

In the learnable adaptation setting, the target VLM backbone remains frozen and only the injected vector parameters $\theta_l$ are optimized. The learnable vectors are initialized from the corresponding extracted Reasoning Vectors unless otherwise specified. We use the same offline support set as vector extraction and optimize only the alignment objective described in the main text; no language-model cross-entropy loss is used for vector adaptation.

We train the vector parameters for 6 epochs using AdamW with learning rate $1\times10^{-4}$. The batch size is 1, with gradient accumulation over 4 steps. We use weight decay $1\times10^{-3}$, warmup ratio 0.1, gradient clipping with maximum norm 1.0, and 16-bit mixed precision. The VLM backbone and vision encoder remain frozen throughout adaptation. Table~\ref{tab:app_learnable_hparams} summarizes the hyperparameters used for learnable vector adaptation.

\begin{table}[htbp]
\centering
\small
\caption{Hyperparameters for learnable vector adaptation.}
\label{tab:app_learnable_hparams}
\begin{tabular}{lc}
\toprule
Hyperparameter & Value \\
\midrule
Trainable parameters & Injected vectors $\theta_l$ only \\
Backbone parameters & Frozen \\
Optimizer & AdamW \\
Learning rate & $1\times10^{-4}$ \\
Batch size & 1 \\
Gradient accumulation & 4 \\
Training epochs & 6 \\
Weight decay & $1\times10^{-3}$ \\
Warmup ratio & 0.1 \\
Gradient clipping & 1.0 \\
Training objective & Alignment loss only \\
Precision & 16-mixed \\
Initialization & LLM-derived or VLM-derived vector \\
\bottomrule
\end{tabular}
\end{table}

\FloatBarrier

\subsection{Reproducibility Details}
\label{app:reproducibility_details}

We use random seed 42 for support-set sampling, layer-selection subset sampling, and learnable adaptation unless otherwise specified. All baseline and LIFT evaluations use deterministic decoding with sampling disabled. Experiments are implemented in PyTorch and run with \texttt{torch.float16} for static evaluation and 16-bit mixed precision for learnable adaptation. We run experiments on NVIDIA RTX 3090 GPUs with 24GB memory and NVIDIA RTX A6000 GPUs with 48GB memory. All experiments fit on a single GPU. The wall-clock time varies across benchmarks depending on the evaluation-set size and maximum generation length: text-only benchmarks with a 512-token limit are faster, while multimodal benchmarks with a 4096-token limit require longer evaluation.

\FloatBarrier

\section{Statistical Uncertainty of Main Accuracy Differences}
\label{app:bootstrap_ci}

We report paired bootstrap confidence intervals to quantify the statistical uncertainty of the main accuracy differences.
For each comparison, we use per-example correctness pairs from the same evaluation split and compute the paired accuracy difference under bootstrap resampling.
We use 10,000 bootstrap resamples for each comparison.

\begin{table}[htbp]
\centering
\small
\caption{Evaluation split sizes used in our experiments.}
\label{tab:eval_sizes}
\begin{tabular}{lc}
\toprule
Benchmark & \#Examples \\
\midrule
GSM8K & 1319 \\
MathVista & 1000 \\
MathVision & 304 \\
CommonsenseQA & 1221 \\
StrategyQA & 687 \\
ScienceQA & 4241 \\
\bottomrule
\end{tabular}
\end{table}

For macro-average comparisons, we first compute the paired accuracy difference within each model--benchmark pair and then average the differences equally across all evaluated model--benchmark pairs. All deltas are reported in percentage points.

\begin{table}[htbp]
\centering
\small
\caption{Paired bootstrap confidence intervals for macro-average accuracy differences.}
\label{tab:bootstrap_macro}
\begin{tabular}{lccc}
\toprule
Comparison & Avg. Delta & 95\% CI \\
\midrule
LLM-derived $-$ Baseline & +1.45 & [0.45, 2.45] \\
VLM-derived $-$ Baseline & +0.70 & [0.12, 1.28] \\
LLM-derived $-$ VLM-derived & +0.75 & [0.15, 1.35] \\
\bottomrule
\end{tabular}
\end{table}

% \begin{table}[htbp]
% \centering
% \small
% \caption{Paired bootstrap confidence intervals for macro-average accuracy differences.}
% \label{tab:bootstrap_macro}
% \begin{tabular}{lccc}
% \toprule
% Comparison & Avg. Delta & 95\% CI & $p_{\mathrm{boot}}$ \\
% \midrule
% LLM-derived $-$ Baseline & +1.45 & [0.45, 2.45] & 0.0030 \\
% VLM-derived $-$ Baseline & +0.70 & [0.12, 1.28] & 0.0212 \\
% LLM-derived $-$ VLM-derived & +0.75 & [0.15, 1.35] & 0.0070 \\
% \bottomrule
% \end{tabular}
% \end{table}

Table~\ref{tab:bootstrap_macro} shows that both LLM-derived and VLM-derived Reasoning Vectors improve macro-average accuracy over the baseline. LLM-derived vectors also achieve higher macro-average accuracy than VLM-derived vectors under this paired bootstrap analysis. 

\FloatBarrier

\section{Robustness Checks and Random-vector Control}
\label{app:robustness_controls}

% \subsection{Additional-seed Robustness Check}
% \label{app:seed_robustness}
% We additionally repeat the experiment with one different random seed for support-set construction and layer-selection development examples, while keeping the evaluation split fixed. Table~\ref{tab:additional_seed} shows that the main trend remains unchanged: LLM-derived Reasoning Vectors improve over the baseline and achieve higher average accuracy than VLM-derived vectors on both VLM backbones.

% \begin{table}[htbp]
% \centering
% \small
% \caption{Additional-seed robustness check for frozen Reasoning Vector transfer.}
% \label{tab:additional_seed}
% \resizebox{\linewidth}{!}{
% \begin{tabular}{llccccccc}
% \toprule
% Model & Method & GSM8K & MathVista & MathVision & CSQA & SQA & SciQA & Avg. \\
% \midrule
% \multirow{3}{*}{Qwen2.5-VL}
% & Baseline & 82.2 & 68.5 & 25.0 & 77.3 & 58.6 & 87.8 & 66.6 \\
% & LLM-derived & \textbf{84.4} & 69.5 & \textbf{26.7} & \textbf{78.3} & \textbf{63.4} & \textbf{88.0} & \textbf{68.4} \\
% & VLM-derived & 83.2 & \textbf{70.6} & 25.7 & 77.8 & 60.3 & 87.6 & 67.5 \\
% \midrule
% \multirow{3}{*}{InternVL2.5}
% & Baseline & 77.1 & 62.8 & 22.7 & 82.3 & 65.6 & 96.5 & 67.8 \\
% & LLM-derived & \textbf{78.5} & \textbf{63.9} & \textbf{25.0} & \textbf{83.5} & \textbf{67.2} & \textbf{96.8} & \textbf{69.2} \\
% & VLM-derived & 77.5 & 63.1 & 24.3 & 82.9 & 65.8 & 96.7 & 68.4 \\
% \bottomrule
% \end{tabular}
% }
% \end{table}

\subsection{Additional-seed Robustness Check}

We additionally repeat the experiment with one different random
seed for support-set construction and layer-selection development
examples, while keeping the evaluation split fixed. Table~\ref{tab:additional_seed}
shows that the main trend remains unchanged: LLM-derived
Reasoning Vectors improve over the baseline and achieve higher
average accuracy than VLM-derived vectors on both VLM backbones.

\begin{table}[t]
\centering
\small
\caption{Additional-seed robustness check for frozen Reasoning
Vector transfer. All results are reported as accuracy (\%).}
\label{tab:additional_seed}
\begin{tabular}{llrrrrrrr}
\toprule
Model & Method & GSM8K & MVista & MVision & CSQA & SQA & SciQA & Avg. \\
\midrule
\multirow{3}{*}{Qwen2.5-VL}
& Baseline    & 82.2 & 68.5 & 25.0 & 77.3 & 58.6 & 87.8 & 66.6 \\
& LLM-derived & 84.4 & 69.5 & 26.7 & 78.3 & 63.4 & 88.0 & 68.4 \\
& VLM-derived & 83.2 & 70.6 & 25.7 & 77.8 & 60.3 & 87.6 & 67.5 \\
\midrule
\multirow{3}{*}{InternVL2.5}
& Baseline    & 77.1 & 62.8 & 22.7 & 82.3 & 65.6 & 96.5 & 67.8 \\
& LLM-derived & 78.5 & 63.9 & 25.0 & 83.5 & 67.2 & 96.8 & 69.2 \\
& VLM-derived & 77.5 & 63.1 & 24.3 & 82.9 & 65.8 & 96.7 & 68.4 \\
\bottomrule
\end{tabular}
\end{table}

We further evaluate the robustness of LIFT on Qwen2.5-VL-7B-Instruct
using three random seeds (42, 347, and 125). In each run, the
support examples and held-out subsets used for layer selection
are independently sampled. We report the mean and standard
deviation across the three runs in Table~\ref{tab:three_seed}.

\begin{table}[t]
\centering
\small
\caption{Three-seed robustness evaluation on Qwen2.5-VL-7B-Instruct.
Results are reported as mean $\pm$ standard deviation of accuracy (\%).}
\label{tab:three_seed}
\begin{tabular}{llrrrrrr}
\toprule
Model & Method & GSM8K & MVista & MVision & CSQA & SQA & SciQA \\
\midrule
\multirow{2}{*}{Qwen2.5-VL}
& LLM-derived
& $84.3 \pm 0.8$ & $69.9 \pm 0.7$ & $26.5 \pm 0.7$
& $78.4 \pm 0.4$ & $62.5 \pm 0.9$ & $88.4 \pm 0.3$ \\
& VLM-derived
& $82.5 \pm 0.5$ & $69.6 \pm 1.0$ & $25.0 \pm 0.7$
& $77.8 \pm 0.4$ & $60.3 \pm 0.5$ & $87.5 \pm 0.5$ \\
\bottomrule
\end{tabular}
\end{table}

Across the three runs, LLM-derived vectors consistently
outperform VLM-derived vectors on all six benchmarks.
The relatively small standard deviations further demonstrate
that the performance improvements are robust to variations
in support-set sampling and layer selection.

\FloatBarrier

\subsection{Same-norm Random-vector Control}
\label{app:same_norm_random}

We further conduct a same-norm random-vector control to test whether the gains of LIFT can be explained by arbitrary activation perturbations.
For each selected intervention layer $l$, we sample a Gaussian random vector $\epsilon_l \sim \mathcal{N}(0,I)$ and rescale it to have the same $\ell_2$ norm as the corresponding LLM-derived Reasoning Vector $r_l$:
\begin{equation}
r_l^{\mathrm{rand}}
=
\frac{\epsilon_l}{\|\epsilon_l\|_2}
\cdot
\|r_l\|_2 .
\end{equation}
We then inject the same-norm random vector into the target VLM using the same selected layer, scaling coefficient, decoding setting, and evaluation protocol as the corresponding LIFT intervention:
\begin{equation}
\widetilde{x}_l
=
x_l + \mu r_l^{\mathrm{rand}} .
\end{equation}

This control preserves the perturbation magnitude while replacing the extracted Reasoning Vector direction with a random direction.
It therefore tests whether the observed improvements are caused merely by adding a vector of similar scale to the hidden states.

As shown in Table~\ref{tab:same_norm_random}, same-norm random vectors do not improve over the baseline and fail to reproduce the gains of LLM-derived Reasoning Vectors.
On Qwen2.5-VL, the random-vector control obtains an average accuracy of 66.4, compared with 66.6 for the baseline and 68.3 for LLM-derived vectors.
On InternVL2.5, it obtains 66.1, below both the baseline average of 67.8 and the LLM-derived average of 69.0.
These results suggest that LIFT's gains are not explained by perturbation magnitude alone; the Reasoning Vectors provide useful task-relevant information for intervention.

\begin{table}[htbp]
\centering
\small
\caption{
Same-norm random-vector control. All results are reported as accuracy (\%).
}
\label{tab:same_norm_random}
\resizebox{\linewidth}{!}{
\begin{tabular}{llccccccc}
\toprule
Model & Method & GSM8K & MathVista & MathVision & CSQA & SQA & SciQA & Avg. \\
\midrule
\multirow{3}{*}{Qwen2.5-VL}
& Baseline & 82.2 & 68.5 & 25.0 & 77.3 & 58.6 & 87.8 & 66.6 \\
& Same-norm random & 82.1 & 68.2 & 25.0 & 77.3 & 58.2 & 87.8 & 66.4 \\
& LLM-derived & \textbf{84.2} & \textbf{69.1} & \textbf{26.6} & \textbf{78.0} & \textbf{63.5} & \textbf{88.1} & \textbf{68.3} \\
\midrule
\multirow{3}{*}{InternVL2.5}
& Baseline & 77.1 & 62.8 & 22.7 & 82.3 & 65.6 & 96.5 & 67.8 \\
& Same-norm random & 77.1 & 62.6 & 22.8 & 79.7 & 58.4 & 96.3 & 66.1 \\
& LLM-derived & \textbf{78.1} & \textbf{64.0} & \textbf{25.0} & \textbf{83.1} & \textbf{67.1} & \textbf{96.8} & \textbf{69.0} \\
\bottomrule
\end{tabular}
}
\end{table}

\subsection{Reliability of Held-out Layer Selection}
\label{app:layer_selection_oracle}

To evaluate the reliability of our layer-selection strategy,
we compare the layer selected using a 100-example held-out
subset with the oracle layer identified through an exhaustive
sweep over candidate intervention layers. As shown in Table~\ref{tab:layer_selection_oracle}, the
performance gap between the selected layer and the oracle
layer is at most 0.3 percentage points across all four settings.
These results demonstrate that our lightweight layer-selection
procedure provides a close approximation to exhaustive layer
search, without requiring evaluation on all candidate layers.

\begin{table}[t]
\centering
\small
\caption{Comparison between held-out selected layers and oracle
layers. Results are averaged over six benchmarks.}
\label{tab:layer_selection_oracle}
\begin{tabular}{llccc}
\toprule
Model & Method & Selected & Oracle & Gap \\
\midrule
\multirow{2}{*}{Qwen2.5-VL}
& LLM-derived & 68.3 & 68.3 & 0.0 \\
& VLM-derived & 67.5 & 67.8 & 0.3 \\
\midrule
\multirow{2}{*}{InternVL2.5}
& LLM-derived & 69.0 & 69.1 & 0.1 \\
& VLM-derived & 68.3 & 68.4 & 0.1 \\
\bottomrule
\end{tabular}
\end{table}

\subsection{Sensitivity to Support-set Size}
\label{app:support_set_size}

We further investigate the sensitivity of LIFT to the number
of support examples by evaluating support-set sizes of 100,
200, and 500. As shown in Table~\ref{tab:support_set_size}, LIFT remains
effective with substantially fewer support examples.
Using only 100 examples already improves performance over
the baseline across all reported settings. These results suggest that LIFT can
extract useful Reasoning Vectors from relatively small
support sets.

\begin{table}[t]
\centering
\small
\caption{Sensitivity to support-set size. Results are averaged over CSQA, SQA, and SciQA.}
\label{tab:support_set_size}
\begin{tabular}{llcccc}
\toprule
Model & Method & Baseline & 100 & 200 & 500 \\
\midrule
\multirow{2}{*}{Qwen2.5-VL}
& LLM-derived & 74.6 & 75.7 & 75.9 & 76.5 \\
& VLM-derived & 74.6 & 75.1 & 75.3 & 75.3 \\
\midrule
\multirow{2}{*}{InternVL2.5}
& LLM-derived & 81.5 & 81.9 & 82.1 & 82.3 \\
& VLM-derived & 81.5 & 81.9 & 81.8 & 81.7 \\
\bottomrule
\end{tabular}
\end{table}

% \FloatBarrier

\section{Additional Attention-level Injection Results}
\label{app:static_injection_variants}

To test whether LIFT can also work at attention-head outputs, we additionally evaluate attention-level injection. This variant extracts head-wise Reasoning Vectors from attention outputs and injects them into the corresponding attention-head outputs during inference. The target VLM backbone remains frozen, and the vision encoder is not modified. For each dataset, the injection layer is selected on a held-out development subset of 100 training examples and then fixed for final evaluation. No evaluation examples are used for layer selection.

For each attention head \(m\) at layer \(l\), the attention-level vector is computed as
\begin{equation}
v_{\mathrm{Attn}}^{(l,m)}
=
\frac{1}{N}\sum_{i=1}^{N}
\frac{1}{|A_i|}
\sum_{a\in A_i}
\left(
o_{R,i}^{(l,m)}(a) - o_{S,i}^{(l,m)}(a)
\right),
\end{equation}
where \(o_{R,i}^{(l,m)}(a)\) and \(o_{S,i}^{(l,m)}(a)\) denote the output of attention head \(m\) for answer token \(a\) under the Reasoner and Solver inputs, respectively.

During inference, the intervention is applied to the attention-head output at each generated language-token position \(t\):
\begin{equation}
\widetilde{o}^{(l,m)}(t)
=
o^{(l,m)}(t)
+
\mu \cdot v_{\mathrm{Attn}}^{(l,m)} .
\end{equation}
Here, \(o^{(l,m)}(t)\) is the original output of attention head \(m\) at layer \(l\), \(\widetilde{o}^{(l,m)}(t)\) is the intervened output, and \(\mu\) is the intervention strength. We set \(\mu=1.0\) for all attention-level injection experiments.

Table~\ref{tab:app_attention_static} reports the results of attention-level injection. Attention-level vectors improve over the baseline on average for both VLM backbones. LLM-derived vectors again achieve higher average performance than VLM-derived vectors. This provides additional evidence that the advantage of LLM-derived vectors is not specific to the hidden-state injection setting used in the main experiments, but can also appear when the intervention is applied to attention-head outputs.

\begin{table}[htbp]
\centering
\caption{
Additional results of Attention-level Injection. For each dataset, the injection layer is selected on a held-out development subset and then fixed for evaluation.
}
\label{tab:app_attention_static}
\resizebox{\linewidth}{!}{
\begin{tabular}{llccccccc}
\toprule
Model & Method & GSM8K & MVista & MVision & CSQA & SQA & SciQA & Avg. \\
\midrule
\multirow{3}{*}{Qwen2.5-VL}
& Baseline & 82.2 & 68.5 & 25.0 & 77.3 & 58.6 & 87.8 & 66.6 \\
& LLM-derived & \textbf{83.2} & \textbf{69.6} & \textbf{26.6} & \textbf{77.8} & \textbf{59.7} & \textbf{88.1} & \textbf{67.5} \\
& VLM-derived & 82.9 & 69.5 & 26.3 & 77.6 & 59.4 & 88.0 & 67.3 \\
\midrule
\multirow{3}{*}{InternVL2.5}
& Baseline & 77.1 & 62.8 & 22.7 & 82.3 & 65.6 & 96.5 & 67.8 \\
& LLM-derived & \textbf{78.1} & \textbf{63.9} & \textbf{25.3} & \textbf{83.2} & \textbf{67.4} & \textbf{96.7} & \textbf{69.1} \\
& VLM-derived & 77.9 & 63.2 & 25.0 & 83.1 & 67.0 & 96.6 & 68.8 \\
\bottomrule
\end{tabular}
}
\end{table}

\FloatBarrier

\section{Additional Case Studies with Full Reasoning Traces}
\label{app:case_study}

We provide additional case studies with full reasoning traces. These examples complement the summarized cases in the main text by showing the outputs of the baseline, VLM-derived Reasoning Vector intervention, and LLM-derived Reasoning Vector intervention under the same zero-shot CoT prompt.

\small

% ---------------- Case 1 ----------------
\begin{caseheader}{Case 1: Target-score}
\textbf{Question:} By shooting two arrows at the shown target on the wall, how many different scores can we obtain?

\textbf{Gold Answer:} 9.
\end{caseheader}

\begin{traceblock}{RVBase}{Baseline}{\textcolor{red}{7} \xmark}
To determine the number of different scores that can be obtained by shooting two arrows at the given target, we need to consider all possible combinations of the numbers in the circles. The target has three concentric circles with the following numbers: - The innermost circle (center): 6 - The middle circle: 3 - The outermost circle: 2 and 0 We will calculate the score for each possible combination of two arrows: 1. Both arrows hit the center (6): - Score = \(6 + 6 = 12\) 2. One arrow hits the center (6) and the other hits the middle (3): - Scores = \(6 + 3 = 9\) or \(3 + 6 = 9\) - Total unique score = 9 3. One arrow hits the center (6) and the other hits the outer circle (2 or 0): - Scores = \(6 + 2 = 8\) or \(6 + 0 = 6\) - Total unique scores = 8 and 6 4. Both arrows hit the middle circle (3): - Score = \(3 + 3 = 6\) 5. One arrow hits the middle circle (3) and the other hits the outer circle (2 or 0): - Scores = \(3 + 2 = 5\) or \(3 + 0 = 3\) - Total unique scores = 5 and 3 6. Both arrows hit the outer circle (2 and 0): - Scores = \(2 + 0 = 2\) or \(0 + 2 = 2\) - Total unique score = 2 Now, let's list all the unique scores: - 12 - 9 - 8 - 6 - 5 - 3 - 2 Counting these unique scores, we have: 12, 9, 8, 6, 5, 3, 2. There are 7 unique scores. Thus, the number of different scores that can be obtained is: \boxed{7}
\end{traceblock}

\begin{traceblock}{RVVLM}{VLM-derived}{\textcolor{red}{10} \xmark}
To determine the number of different scores that can be obtained by shooting two arrows at the given target, we need to consider all possible combinations of the numbers in the circles. The target has three concentric circles with the following numbers: - The innermost circle (center): 6 - The middle circle: 3 - The outermost circle: 2 and 0 We can shoot an arrow into any of these circles, so there are four possible outcomes for each arrow. Let's list out all the possible combinations: 1. Both arrows hit the center (6) 2. One arrow hits the center (6) and the other hits the middle (3) 3. One arrow hits the center (6) and the other hits the outer (2 or 0) 4. Both arrows hit the middle (3) 5. One arrow hits the middle (3) and the other hits the outer (2 or 0) 6. Both arrows hit the outer (2 or 0) Now, let's count the unique scores: 1. 6 + 6 = 12 2. 6 + 3 = 9 3. 6 + 2 = 8 4. 6 + 0 = 6 5. 3 + 3 = 6 6. 3 + 2 = 5 7. 3 + 0 = 3 8. 2 + 2 = 4 9. 2 + 0 = 2 10. 0 + 0 = 0 There are 10 different scores that can be obtained by shooting two arrows at the target. \boxed{10}
\end{traceblock}

\begin{traceblock}{RVLLM}{LLM-derived}{\textcolor{RVLLM}{9} \cmark}
To determine the number of different scores that can be obtained by shooting two arrows at the given target, we need to consider all possible combinations of the numbers in the circles. The target has three concentric circles with the following numbers: - The innermost circle contains the number 6. - The middle circle contains the number 3. - The outermost circle contains the numbers 2 and 0. We will calculate the score for each possible combination of two arrows: 1. Combination (6, 6): - Score = \(6 + 6 = 12\) 2. Combination (6, 3): - Score = \(6 + 3 = 9\) 3. Combination (6, 2): - Score = \(6 + 2 = 8\) 4. Combination (6, 0): - Score = \(6 + 0 = 6\) 5. Combination (3, 3): - Score = \(3 + 3 = 6\) 6. Combination (3, 2): - Score = \(3 + 2 = 5\) 7. Combination (3, 0): - Score = \(3 + 0 = 3\) 8. Combination (2, 2): - Score = \(2 + 2 = 4\) 9. Combination (2, 0): - Score = \(2 + 0 = 2\) 10. Combination (0, 0): - Score = \(0 + 0 = 0\) Now, let's list all the unique scores obtained from these combinations: 12, 9, 8, 6, 5, 3, 4, 2, 0. Counting the unique scores, we have: 9 \text{ unique scores}. Thus, the number of different scores that can be obtained is: \boxed{9}
\end{traceblock}

\noindent\textbf{Observation.} The LLM-derived vector repairs both repeated-hit enumeration and score deduplication, while the VLM-derived vector enumerates more outcomes but still counts pair outcomes rather than distinct scores.

% ---------------- Case 2 ----------------
\begin{caseheader}{Case 2: Gear}
\textbf{Question:} Four cogs are connected to each other as shown in the picture. The first has 30 teeth, the second 15, the third 60 and the fourth 10. How many turns will the last cog make for each full turn of the first cog?

\textbf{Gold Answer:} 3.
\end{caseheader}

\begin{traceblock}{RVBase}{Baseline}{\textcolor{red}{6} \xmark}
To solve this problem, we need to understand how the number of teeth on each gear affects the rotation of the next gear. 1. First Gear (30 teeth): - When it makes one full turn, it will rotate 30 teeth. 2. Second Gear (15 teeth): - Since the first gear has 30 teeth and the second gear has 15 teeth, the second gear will rotate twice for every one full turn of the first gear. - Therefore, for each full turn of the first gear, the second gear will make 2 turns. 3. Third Gear (60 teeth): - The third gear has 60 teeth. For every full turn of the second gear (which is 2 turns), the third gear will rotate half a turn because \(60 / 2 = 30\). - So, for each full turn of the first gear, the third gear will make 1 turn. 4. Fourth Gear (10 teeth): - The fourth gear has 10 teeth. For every full turn of the third gear (which is 1 turn), the fourth gear will rotate six times because \(10 / 1 = 10\). - Therefore, for each full turn of the first gear, the fourth gear will make 6 turns. So, the final answer is: \boxed{6} 
\end{traceblock}

\begin{traceblock}{RVVLM}{VLM-derived}{\textcolor{red}{6} \xmark}
To determine how many turns the last cog makes for each full turn of the first cog, we need to calculate the gear ratio between the first and the last cog. The gear ratio is given by the number of teeth on one gear divided by the number of teeth on the gear it is connected to. For the first and second cogs: - The first cog has 30 teeth. - The second cog has 15 teeth. - Gear ratio = \( \frac{30}{15} = 2 \). This means that for every full turn of the first cog, the second cog will make 2 turns. Next, for the second and third cogs: - The second cog has 15 teeth. - The third cog has 60 teeth. - Gear ratio = \( \frac{60}{15} = 4 \). This means that for every full turn of the second cog (which is equivalent to 2 turns of the first cog), the third cog will make 4 turns. Finally, for the third and fourth cogs: - The third cog has 60 teeth. - The fourth cog has 10 teeth. - Gear ratio = \( \frac{60}{10} = 6 \). This means that for every full turn of the third cog (which is equivalent to 8 turns of the first cog), the fourth cog will make 6 turns. Therefore, for each full turn of the first cog, the last cog (fourth cog) will make 6 turns. \boxed{6}
\end{traceblock}

\begin{traceblock}{RVLLM}{LLM-derived}{\textcolor{RVLLM}{3} \cmark}
To determine how many turns the last cog makes for each full turn of the first cog, we need to calculate the gear ratio between the first and the last cog. The gear ratio is given by the formula: $ \text{Gear Ratio} = \frac{\text{Number of Teeth on First Cog}}{\text{Number of Teeth on Last Cog}} $ Given: - Number of Teeth on First Cog (T1) = 30 - Number of Teeth on Last Cog (T4) = 10 Using the formula: $ \text{Gear Ratio} = \frac{30}{10} = 3 $ This means that for every full turn of the first cog, the last cog will make 3 turns. Therefore, the final answer is: $ \boxed{3} $
\end{traceblock}

\noindent\textbf{Observation.} The LLM-derived vector preserves the end-to-end gear relation, while the baseline and VLM-derived vector remain dominated by local ratio errors.

% ---------------- Case 3 ----------------
\begin{caseheader}{Case 3: Frog-counting}
\textbf{Question:} In the diagram you can see a very ragged island. Some of the frogs are sitting in the water. How many are sitting on the island?

\textbf{Gold Answer:} 6.
\end{caseheader}

\begin{traceblock}{RVBase}{Baseline}{\textcolor{red}{3} \xmark}
To solve this problem, we need to count the number of frogs that are sitting on the island, not in the water. 1. First, let's identify the frogs: - There are 10 frogs in total. - 7 frogs are in the water (as indicated by their position outside the landmass). - The remaining frogs must be on the island. 2. Now, subtract the number of frogs in the water from the total number of frogs: - Total frogs = 10 - Frogs in water = 7 - Frogs on the island = Total frogs - Frogs in water - Frogs on the island = 10 - 7 = 3 Therefore, the number of frogs sitting on the island is $\boxed{3}$.
\end{traceblock}

\begin{traceblock}{RVVLM}{VLM-derived}{\textcolor{red}{3} \xmark}
To determine how many frogs are sitting on the island, let's follow these steps: 1. Count all the frogs in the image. 2. Identify which frogs are on the island. Step 1: Counting all the frogs: - There are 9 frogs in total in the image. Step 2: Identifying frogs on the island: - The island is the area with the palm tree. - Upon inspection, there are 3 frogs on the island. Therefore, the number of frogs sitting on the island is $\boxed{3}$.
\end{traceblock}

\begin{traceblock}{RVLLM}{LLM-derived}{\textcolor{RVLLM}{6} \cmark}
To determine how many frogs are sitting on the island, we need to count all the frogs that are not in the water. 1. There is one frog at the top left corner. 2. There are two frogs in the middle section. 3. There is one frog at the bottom right corner. 4. There are two frogs near the palm tree. Adding these up: - Top left: 1 frog - Middle section: 2 frogs - Bottom right: 1 frog - Near the palm tree: 2 frogs Total: \(1 + 2 + 1 + 2 = 6\) So, there are 6 frogs sitting on the island. $\boxed{6}$
\end{traceblock}

\noindent\textbf{Observation.} The LLM-derived vector follows the island boundary by regions, while the VLM-derived vector narrows the target region to the palm-tree area.

% ---------------- Case 4 ----------------
\begin{caseheader}{Case 4: Workout-weeks}
\textbf{Question:} After tracking his workout routine for 8 weeks, Josh discovered that he had spent 4 hours working out every week except on some occasions. At one point, he spent 5 hours each for two consecutive weeks. Another time, he spent 6 hours in one week. What is the total number of hours he has spent working out across the 8 weeks?

\textbf{Gold Answer:} 36.
\end{caseheader}

\begin{traceblock}{RVBase}{Baseline}{\textcolor{red}{40} \xmark}
To determine the total number of hours Josh has spent working out over the 8 weeks, we need to account for the specific weeks where he worked out more than the usual 4 hours. 1. \textbf{Calculate the hours for the usual weeks:} - There are 8 weeks in total. - For 6 weeks, Josh worked out 4 hours each week. $ 6 \text{ weeks} \times 4 \text{ hours/week} = 24 \text{ hours} $ 2. \textbf{Account for the weeks with additional hours:} - Josh worked out 5 hours each for 2 consecutive weeks. $ 2 \text{ weeks} \times 5 \text{ hours/week} = 10 \text{ hours} $ - Josh also worked out 6 hours in one week. $ 1 \text{ week} \times 6 \text{ hours/week} = 6 \text{ hours} $ 3. \textbf{Sum all the hours together:} $ 24 \text{ hours} + 10 \text{ hours} + 6 \text{ hours} = 40 \text{ hours} $ Thus, the total number of hours Josh has spent working out across the 8 weeks is: $\boxed{40}$
\end{traceblock}

\begin{traceblock}{RVVLM}{VLM-derived}{\textcolor{red}{40} \xmark}
To determine the total number of hours Josh has spent working out over the 8 weeks, we need to account for the specific weeks where he worked out more than 4 hours. First, let's identify the weeks where Josh worked out more than 4 hours: - He worked out 5 hours each for two consecutive weeks. - He worked out 6 hours in one week. Now, let's calculate the total hours for these specific weeks: 1. For the two consecutive weeks where he worked out 5 hours each: $ 5 \text{ hours/week} \times 2 \text{ weeks} = 10 \text{ hours} $ 2. For the one week where he worked out 6 hours: $ 6 \text{ hours/week} \times 1 \text{ week} = 6 \text{ hours} $ Next, we add these hours to the total hours he worked out for the remaining 6 weeks (since there are 8 weeks in total): $ 4 \text{ hours/week} \times 6 \text{ weeks} = 24 \text{ hours} $ Finally, we sum all the hours together: $ 24 \text{ hours} + 10 \text{ hours} + 6 \text{ hours} = 40 \text{ hours} $ Thus, the total number of hours Josh has spent working out across the 8 weeks is: $ \boxed{40} $
\end{traceblock}

\begin{traceblock}{RVLLM}{LLM-derived}{\textcolor{RVLLM}{36} \cmark}
To determine the total number of hours Josh has spent working out over the 8 weeks, we need to account for the specific weeks where he worked out more than 4 hours. First, let's identify the weeks where Josh worked out more than 4 hours: - He worked out 5 hours each for two consecutive weeks. - He worked out 6 hours in one week. So, the total hours for these specific weeks can be calculated as follows: 1. For the two consecutive weeks where he worked out 5 hours each: $ 5 \text{ hours/week} \times 2 \text{ weeks} = 10 \text{ hours} $ 2. For the one week where he worked out 6 hours: $ 6 \text{ hours/week} \times 1 \text{ week} = 6 \text{ hours} $ Next, we add these hours together with the hours from the remaining 5 weeks (since there are 8 weeks in total): $ 4 \text{ hours/week} \times 5 \text{ weeks} = 20 \text{ hours} $ Now, summing all the hours: $ 20 \text{ hours} + 10 \text{ hours} + 6 \text{ hours} = 36 \text{ hours} $ Therefore, the total number of hours Josh has spent working out across the 8 weeks is: $ \boxed{36} $
\end{traceblock}

\noindent\textbf{Observation.} The LLM-derived vector updates the remaining-week count, while the baseline and VLM-derived vector double-count the special weeks.

% ---------------- Case 5 ----------------
\begin{caseheader}{Case 5: Beach-week}
\textbf{Question:} Conor lives near a beach and loves going there every day to have fun. On a particular week, he found 50 people at the beach on the first day. The next day, 20 more people were present at the beach than on the first day, and on the third day, the total number of people increased to twice the number that were there on the second day. If Conor saw an average of 60 people each day the rest of the week, calculate the total number of people Conor saw at the beach that week.

\textbf{Gold Answer:} 500.
\end{caseheader}

\begin{traceblock}{RVBase}{Baseline}{\textcolor{red}{460} \xmark}
To determine the total number of people Conor saw at the beach over the week, we need to calculate the number of people seen each day and then sum these amounts. 1. \textbf{First Day:} - Number of people: \(50\) 2. \textbf{Second Day:} - Number of people: \(50 + 20 = 70\) 3. \textbf{Third Day:} - Number of people: \(2 \times 70 = 140\) 4. \textbf{Rest of the Week (Days 4 through 7):} - Average number of people per day: \(60\) - Number of days in this period: \(7 - 3 = 4\) days - Total number of people for the rest of the week: \(60 \times 4 = 240\) Now, we sum the number of people seen each day: $ 50 + 70 + 140 + 240 = 460 $ Thus, the total number of people Conor saw at the beach that week is \( \boxed{460} \).
\end{traceblock}

\begin{traceblock}{RVVLM}{VLM-derived}{\textcolor{red}{460} \xmark}
To determine the total number of people Conor saw at the beach that week, we need to calculate the number of people seen each day and then sum them up. 1. \textbf{First Day:} - Number of people: \(50\) 2. \textbf{Second Day:} - Number of people: \(50 + 20 = 70\) 3. \textbf{Third Day:} - Number of people: \(2 \times 70 = 140\) 4. \textbf{Rest of the Week (Days 4 through 7):} - Average number of people per day: \(60\) - Number of days in this period: \(7 - 3 = 4\) days - Total number of people for these 4 days: \(60 \times 4 = 240\) Now, we sum up the number of people seen over all seven days: $ 50 + 70 + 140 + 240 = 460 $ Thus, the total number of people Conor saw at the beach that week is \( \boxed{460} \).
\end{traceblock}

\begin{traceblock}{RVLLM}{LLM-derived}{\textcolor{RVLLM}{500} \cmark}
To determine the total number of people Conor saw at the beach that week, we need to calculate the number of people seen each day and then sum these amounts. 1. \textbf{First Day:} - Number of people: \(50\) 2. \textbf{Second Day:} - Number of people: \(50 + 20 = 70\) 3. \textbf{Third Day:} - Number of people: \(2 \times 70 = 140\) 4. \textbf{Rest of the Week (Days 4 through 7):} - Average number of people per day: \(60\) - Number of days in this period: \(7 - 3 = 4\) days - Total number of people for the rest of the week: \(60 \times 4 = 240\) Now, we sum the number of people seen over all seven days: $ 50 + 70 + 140 + 240 = 500 $ Thus, the total number of people Conor saw at the beach that week is \( \boxed{500} \).
\end{traceblock}

\noindent\textbf{Observation.} The LLM-derived vector preserves the arithmetic state after setting up the correct weekly decomposition.

% ---------------- Case 6 ----------------
\begin{caseheader}{Case 6: Marble-price}
\textbf{Question:} If a bag of marbles costs \$20 and the price increases by 20\% of the original price every two months, how much would a bag of marbles cost after 36 months?

\textbf{Gold Answer:} 92.
\end{caseheader}

\begin{traceblock}{RVBase}{Baseline}{\textcolor{red}{80} \xmark}
To determine the cost of a bag of marbles after 36 months, given that the price increases by 20\% every two months, we can follow these steps: 1. \textbf{Identify the initial price}: The initial price of the bag of marbles is \$20. 2. \textbf{Determine the number of price increases over 36 months}: - Since the price increases every two months, we calculate the number of 2-month periods in 36 months. $ \frac{36 \text{ months}}{2 \text{ months/period}} = 18 \text{ periods} $ 3. \textbf{Calculate the total percentage increase over 36 months}: - Each period (2 months) the price increases by 20\%. Over 18 periods, the total increase is: $ 18 \times 20\% = 360\% $ - This means the price will be multiplied by \(1 + 360\% = 4\) times the original price. 4. \textbf{Compute the final price}: - The final price after 36 months is: $ 20 \times 4 = 80 $ Therefore, the cost of a bag of marbles after 36 months is \( \boxed{80} \).
\end{traceblock}

\begin{traceblock}{RVVLM}{VLM-derived}{\textcolor{red}{358.17} \xmark}
To determine the cost of a bag of marbles after 36 months, given that the price increases by 20\% every two months, we can follow these steps: 1. \textbf{Identify the initial price}: The initial price of the bag of marbles is \$20. 2. \textbf{Determine the number of price increases over 36 months}: Since the price increases every two months, we calculate the number of price increases in 36 months: $ \frac{36 \text{ months}}{2 \text{ months per increase}} = 18 \text{ increases} $ 3. \textbf{Calculate the multiplier for each price increase}: Each price increase is a 20\% increase, which can be represented as a multiplication factor of \(1 + 0.20 = 1.20\). 4. \textbf{Apply the multiplier iteratively}: We need to apply this multiplier 18 times to the initial price. This can be expressed as: $ \text{Final Price} = 20 \times (1.20)^{18} $ 5. \textbf{Compute the final price}: $ (1.20)^{18} \approx 17.90847696 $ $ \text{Final Price} = 20 \times 17.90847696 \approx 358.17 $ Therefore, the cost of a bag of marbles after 36 months is approximately \( \boxed{358.17} \).
\end{traceblock}

\begin{traceblock}{RVLLM}{LLM-derived}{\textcolor{RVLLM}{92} \cmark}
To determine the cost of a bag of marbles after 36 months with a price increase of 20\% every two months, we can follow these steps: 1. \textbf{Identify the initial price}: The initial price of the bag of marbles is \$20. 2. \textbf{Determine the number of price increases}: Since the price increases every two months, we need to find out how many times this happens in 36 months. $ \text{Number of price increases} = \frac{36 \text{ months}}{2 \text{ months per increase}} = 18 $ 3. \textbf{Calculate the total percentage increase over 36 months}: Each price increase is 20\%, so after 18 increases, the total percentage increase is: $ \text{Total percentage increase} = 18 \times 20\% = 360\% $ This means the final price will be the original price plus 360\% of the original price. 4. \textbf{Express the final price as a multiple of the original price}: $ \text{Final price} = \text{Original price} + 360\% \times \text{Original price} = 1 \times \text{Original price} + 3.6 \times \text{Original price} = 4.6 \times \text{Original price} $ Therefore, the final price is: $ \text{Final price} = 4.6 \times 20 = 92 $ So, the cost of a bag of marbles after 36 months is $\boxed{92}$.
\end{traceblock}

\noindent\textbf{Observation.} The LLM-derived vector keeps the question's linear ``of the original price'' condition, while the VLM-derived vector switches to compounding.

% ---------------- Case 7 ----------------
\begin{caseheader}{Case 7: Object-subtraction}
\textbf{Question:} Subtract all red matte cubes. Subtract all small green metal objects. How many objects are left?

\textbf{Gold Answer:} 7.
\end{caseheader}

\begin{traceblock}{RVBase}{Baseline}{\textcolor{red}{6} \xmark}
To solve the problem, let's follow these steps: 1. Identify all objects in the image. 2. Subtract all red matte cubes. 3. Subtract all small green metal objects. 4. Count the remaining objects. Step 1: Identify all objects in the image. - There is a red matte cube. - There is a gray cylinder. - There is a red metallic sphere. - There is a blue matte cube. - There is a cyan matte cube. - There is a purple metallic sphere. - There is a green metallic sphere. - There is a brown matte sphere. Step 2: Subtract all red matte cubes. - There is one red matte cube. Removing it leaves us with 7 objects. Step 3: Subtract all small green metal objects. - There is one green metallic sphere. Removing it leaves us with 6 objects. Step 4: Count the remaining objects. - After removing the red matte cube and the green metallic sphere, we are left with 6 objects. The final answer is $\boxed{6}$.
\end{traceblock}

\begin{traceblock}{RVVLM}{VLM-derived}{\textcolor{red}{6} \xmark}
To solve the problem, let's follow these steps: 1. Identify all objects in the image. 2. Subtract the red matte cubes. 3. Subtract any small green metal objects. Step 1: Identify all objects in the image. - There is a red matte cube. - There is a gray cylinder. - There is a red metallic sphere. - There is a blue matte cube. - There is a cyan matte cube. - There is a green metallic sphere. - There is a purple metallic sphere. - There is a brown matte sphere. Step 2: Subtract the red matte cubes. - There is 1 red matte cube. Removing it leaves us with 7 objects. Step 3: Subtract any small green metal objects. - There is 1 small green metallic sphere. Removing it leaves us with 6 objects. The final count of objects left is $\boxed{6}$.
\end{traceblock}

\begin{traceblock}{RVLLM}{LLM-derived}{\textcolor{RVLLM}{7} \cmark}
To solve the problem, let's follow these steps: 1. Identify all objects in the image. 2. Subtract all red matte cubes. 3. Subtract all small green metal objects. 4. Count the remaining objects. Step 1: Identify all objects in the image: - There is a red matte cube. - There is a small gray cylinder. - There is a small red metallic sphere. - There is a small blue matte cube. - There is a small brown matte sphere. - There is a large cyan matte cube. - There is a large green metallic sphere. There is a small purple metallic sphere. Step 2: Subtract all red matte cubes: - There is one red matte cube, so we subtract it. - Remaining objects: small gray cylinder, small red metallic sphere, small blue matte cube, small brown matte sphere, large cyan matte cube, large green metallic sphere, small purple metallic sphere. Step 3: Subtract all small green metal objects: - There are no small green metal objects in the image, so nothing is subtracted here. - Remaining objects: small gray cylinder, small red metallic sphere, small blue matte cube, small brown matte sphere, large cyan matte cube, large green metallic sphere, small purple metallic sphere. Step 4: Count the remaining objects: - Small gray cylinder: 1 - Small red metallic sphere: 1 - Small blue matte cube: 1 - Small brown matte sphere: 1 - Large cyan matte cube: 1 - Large green metallic sphere: 1 - Small purple metallic sphere: 1 Total remaining objects: $\boxed{7}$
\end{traceblock}

\noindent\textbf{Observation.} The LLM-derived vector preserves the full attribute condition and avoids deleting an object that does not satisfy ``small green metal.''

% ---------------- Case 8 ----------------
\begin{caseheader}{Case 8: Dark-blue bar}
\textbf{Question:} What is the difference between highest and lowest value of dark blue bar?

\textbf{Gold Answer:} 53.
\end{caseheader}

\begin{traceblock}{RVBase}{Baseline}{\textcolor{red}{50} \xmark}
To find the difference between the highest and lowest values of the dark blue bars, we need to identify these two values from the chart. The highest value of the dark blue bar is for YouTube, which is 72\%. The lowest value of the dark blue bar is for Snapchat, which is 22\%. Now, we subtract the lowest value from the highest value: $ 72\% - 22\% = 50\% $ So, the difference between the highest and lowest value of the dark blue bar is $\boxed{50}$.
\end{traceblock}

\begin{traceblock}{RVVLM}{VLM-derived}{\textcolor{red}{50} \xmark}
To find the difference between the highest and lowest values of the dark blue bars, we need to identify these two values from the chart. The highest value of the dark blue bar is for YouTube, which is 72\%. The lowest value of the dark blue bar is for Snapchat, which is 22\%. Now, we subtract the lowest value from the highest value: $ 72\% - 22\% = 50\% $ So, the difference between the highest and lowest value of the dark blue bar is $\boxed{50}$.
\end{traceblock}

\begin{traceblock}{RVLLM}{LLM-derived}{\textcolor{RVLLM}{53} \cmark}
To find the difference between the highest and lowest values of the dark blue bars (representing rural users), we need to identify these two values from the chart. The highest value for the dark blue bar is 75\% (for Facebook). The lowest value for the dark blue bar is 22\% (for Snapchat). Now, subtract the lowest value from the highest value: $ 75\% - 22\% = 53\% $ So, the difference between the highest and lowest value of the dark blue bar is $\boxed{53}$.
\end{traceblock}

\noindent\textbf{Observation.} The LLM-derived vector leads to a response that is more consistent with the visual evidence.

% ---------------- Case 9 ----------------
\begin{caseheader}{Case 9: Sleepwalking}
\textbf{Question:} Kim sleepwalks, to monitor her sleeping hours, she installs a camera in her room. Kim usually goes to sleep at 10 p.m. and wakes up at 6 a.m. She is a sleepwalker. One day, after reviewing the cameras, she finds that she woke up that day and was sleepwalking from 2:15 to 2:38 a.m. Also, that day she woke up 5 minutes earlier than usual to go to the bathroom. How many minutes did she sleep on her bed that day?

\textbf{Gold Answer:} 452.
\end{caseheader}

\begin{traceblock}{RVBase}{Baseline}{\textcolor{red}{120} \xmark}
To determine how many minutes Kim slept on her bed that day, we need to break down the information given and calculate step by step. 1. \textbf{Calculate Kim's usual sleeping hours:} - Kim usually goes to sleep at 10 p.m. - Kim usually wakes up at 6 a.m. - The total number of hours Kim sleeps each night is \(6 \text{ a.m.} - 10 \text{ p.m.} = 18\) hours. 2. \textbf{Convert the usual sleeping hours into minutes:} - \(18 \text{ hours} \times 60 \text{ minutes/hour} = 1080 \text{ minutes}\). 3. \textbf{Determine the time Kim woke up that day:} - Kim woke up 5 minutes earlier than usual. - Usual wake-up time is 6 a.m. - Therefore, the actual wake-up time is \(6 \text{ a.m.} - 5 \text{ minutes} = 5:55 \text{ a.m.}\). 4. \textbf{Calculate the time Kim was awake before going to bed:} - Kim went to bed at 10 p.m. - Kim woke up at 5:55 a.m. - The time Kim was awake before going to bed is \(5:55 \text{ a.m.} - 10 \text{ p.m.}\). - Convert 10 p.m. to 22:00 (in 24-hour format). - Calculate the time difference: \(24:00 \text{ (midnight)} - 22:00 \text{ (10 p.m.)} = 2 \text{ hours}\). - Convert 2 hours to minutes: \(2 \text{ hours} \times 60 \text{ minutes/hour} = 120 \text{ minutes}\). - From 10 p.m. to 5:55 a.m., the time Kim was awake is \(120 \text{ minutes} + 5:55 \text{ a.m.} - 10 \text{ p.m.} = 120 \text{ minutes} + 5:55 \text{ a.m.} - 10 \text{ p.m.} = 120\)
\end{traceblock}

\begin{traceblock}{RVVLM}{VLM-derived}{\textcolor{red}{572} \xmark}
To determine how many minutes Kim slept on her bed that day, we need to break down the information given: 1. Kim's usual bedtime is 10 p.m. 2. Kim's usual wake-up time is 6 a.m. 3. On the day in question, she woke up 5 minutes earlier than usual, so her wake-up time was 5:55 a.m. 4. She was sleepwalking from 2:15 a.m. to 2:38 a.m. First, let's calculate the total time Kim spent awake: - From 2:15 a.m. to 2:38 a.m. is \(2:38 - 2:15 = 23\) minutes. Next, let's calculate the total time Kim should have been asleep if she had followed her usual schedule: - From 10 p.m. to 6 a.m. is \(18\) hours, which is \(18 \times 60 = 1080\) minutes. Since she woke up 5 minutes earlier than usual, her actual wake-up time was 5:55 a.m., which means she should have been asleep until 6:05 a.m. (since 5 minutes later would be 6:00 a.m.). Now, let's calculate the time she actually slept: - From 10 p.m. to 5:55 a.m. is \(9\) hours and \(55\) minutes, which is \(9 \times 60 + 55 = 540 + 55 = 595\) minutes. Finally, we subtract the time she was sleepwalking from the total time she should have slept: - Total time she should have slept: 595 minutes - Time she was sleepwalking: 23 minutes So, the time she actually slept on her bed that day is: $ 595 - 23 = 572 \text{ minutes} $ Thus, the number of minutes Kim slept on her bed that day is \( \boxed{572} \).
\end{traceblock}

\begin{traceblock}{RVLLM}{LLM-derived}{\textcolor{RVLLM}{452} \cmark}
To determine how many minutes Kim slept on her bed that day, we need to break down the information given and calculate step by step. 1. \textbf{Normal Sleep Schedule}: - Kim usually goes to sleep at 10 p.m. - Kim usually wakes up at 6 a.m. - Therefore, her normal sleep duration is from 10 p.m. to 6 a.m., which is 8 hours or 480 minutes. 2. \textbf{Sleepwalking Incident}: - On the day in question, Kim woke up 5 minutes earlier than usual to go to the bathroom. - Her usual wake-up time is 6 a.m., so waking up 5 minutes earlier means she woke up at 5:55 a.m. - During this time, she was sleepwalking from 2:15 a.m. to 2:38 a.m. 3. \textbf{Calculate Sleepwalking Duration}: - From 2:15 a.m. to 2:38 a.m. is \(2:38 - 2:15 = 23\) minutes. 4. \textbf{Total Time Kim Was Awake}: - Kim was awake from 5:55 a.m. to 6:00 a.m. (5 minutes) plus the sleepwalking period from 2:15 a.m. to 2:38 a.m. (23 minutes). - Total awake time = 5 minutes + 23 minutes = 28 minutes. 5. \textbf{Calculate Time Spent Sleeping}: - Since Kim was awake for 28 minutes, the time she spent sleeping on her bed that day is the total sleep duration minus the time she was awake. - Total sleep duration = 480 minutes (normal sleep) - 28 minutes (awake time) = 452 minutes. Therefore, the number of minutes Kim slept on her bed that day is \( \boxed{452} \).
\end{traceblock}

\noindent\textbf{Observation.} The LLM-derived vector maintains the overnight interval correctly and subtracts only the awake periods relevant to sleeping on the bed.

% ---------------- Case 10 ----------------
\begin{caseheader}{Case 10: DMV rate-of-change}
\textbf{Question:} An administrator at the Department of Motor Vehicles (DMV) tracked the average wait time from month to month. According to the table, what was the rate of change between August and September? (Unit: minutes per month.)

\textbf{Gold Answer:} \(-3\).
\end{caseheader}

\begin{traceblock}{RVBase}{Baseline}{\textcolor{red}{3} \xmark}
To find the rate of change in waiting time from August to September, we need to calculate the difference in waiting times between these two months and then divide by the number of months between them. The waiting time in August is 17 minutes. The waiting time in September is 14 minutes. The difference in waiting time is: $ 17 - 14 = 3 \text{ minutes} $ Since we are looking at the change over one month, the rate of change is: $ \frac{3 \text{ minutes}}{1 \text{ month}} = 3 \text{ minutes per month} $ Therefore, the rate of change between August and September is $\boxed{3}$.
\end{traceblock}

\begin{traceblock}{RVVLM}{VLM-derived}{\textcolor{red}{3} \xmark}
To find the rate of change between August and September, we need to calculate the difference in waiting times and divide it by the number of months between them. The waiting time in August is 17 minutes. The waiting time in September is 14 minutes. The difference in waiting time is: $ 17 - 14 = 3 \text{ minutes} $ Since there is only one month between August and September, the rate of change is: $ \frac{3 \text{ minutes}}{1 \text{ month}} = 3 \text{ minutes per month} $ So, the rate of change between August and September is $\boxed{3}$.
\end{traceblock}

\begin{traceblock}{RVLLM}{LLM-derived}{\textcolor{RVLLM}{\(-3\)} \cmark}
To find the rate of change in waiting time from August to September, we need to calculate the difference in waiting times between these two months. The waiting time in August is 17 minutes. The waiting time in September is 14 minutes. The rate of change is calculated as follows: Rate of change = Change in waiting time / Number of months Rate of change = (Waiting time in September - Waiting time in August) / (Number of months) Rate of change = (14 - 17) / (September - August) Rate of change = (-3) / 1 Rate of change = -3 minutes per month So, the rate of change between August and September is $\boxed{-3}$ minutes per month.
\end{traceblock}

\noindent\textbf{Observation.} The LLM-derived vector preserves the signed direction of change rather than reporting an unsigned difference.

\normalsize

Overall, these cases show that Reasoning Vector interventions affect the intermediate reasoning process rather than only changing the final answer. In several examples, LLM-derived vectors correct an early reasoning decision and keep the subsequent computation aligned with the question target. VLM-derived vectors can also modify the reasoning trace, but they more often preserve baseline mistakes or drift to a different intermediate target.

% \clearpage
% \input{checklist}

\end{document}